\documentclass[10pt,twocolumn,letterpaper]{article}

\usepackage[pagenumbers]{cvpr} 
\usepackage{balance}
\usepackage{dsfont}

\usepackage{multibib}
\usepackage{etoolbox}
\makeatletter
\providecommand{\sf@counterlist}{}
\pretocmd{\multicols}{\let\sf@counterlist\@empty}{}{}
\makeatother
\newcites{app}{References}
\usepackage{graphicx}
\usepackage{svg}
\usepackage{booktabs}
\usepackage{rotating}
\usepackage{multirow} 
\usepackage{pifont}
\usepackage{textcomp}
\usepackage{stfloats}
\usepackage{url}
\usepackage{verbatim}
\usepackage{amsmath,amsfonts}
\usepackage{algorithmic}
\usepackage{array}
\usepackage[caption=false,font=normalsize,labelfont=sf,textfont=sf]{subfig}

\definecolor{cvprblue}{rgb}{0.21,0.49,0.74}
\usepackage[pagebackref,breaklinks,colorlinks,allcolors=cvprblue]{hyperref}

\def\paperID{9661} 
\def\confName{CVPR}
\def\confYear{2025}

\title{AnomalyNCD: Towards Novel Anomaly Class Discovery in Industrial Scenarios}

\author{
Ziming Huang\textsuperscript{1,$*$} \quad
Xurui Li\textsuperscript{1,$*$} \quad
Haotian Liu\textsuperscript{1,$*$} \quad
Feng Xue\textsuperscript{4} \quad
Yuzhe Wang\textsuperscript{1} \quad
Yu Zhou\textsuperscript{1,2,3,$\dagger$} \\
$^1$School of Electronic Information and Communications,
Huazhong University of Science and Technology \\
$^2$ Hubei Key Laboratory of Smart Internet Technology,
Huazhong University of Science and Technology \\
$^3$Artificial Intelligence Research Institute, Wuhan JingCe Electronic Group Co., LTD \\
$^4$Department of Information Engineering and Computer Science, University of Trento
}

\begin{document}

\maketitle
\begin{abstract}
Recently, multi-class anomaly classification has garnered increasing attention.
Previous methods directly cluster anomalies but often struggle due to the lack of anomaly-prior knowledge.
Acquiring this knowledge faces two issues: the non-prominent and weak-semantics anomalies.
In this paper,
we propose AnomalyNCD,
a multi-class anomaly classification network compatible with different anomaly detection methods.
To address the non-prominence of anomalies,
we design main element binarization (MEBin) to obtain anomaly-centered images,
ensuring anomalies are learned while avoiding the impact of incorrect detections.
Next, to learn anomalies with weak semantics,
we design mask-guided representation learning,
which focuses on isolated anomalies guided by masks
and reduces confusion from erroneous inputs through corrected pseudo labels.
Finally, to enable flexible classification at both region and image levels,
we develop a region merging strategy that determines the overall image category based on the classified anomaly regions.
Our method outperforms the state-of-the-art works on the MVTec AD and MTD datasets.
Compared with the current methods,
AnomalyNCD combined with zero-shot anomaly detection method achieves a 10.8\% $F_1$ gain,
8.8\% NMI gain,
and 9.5\% ARI gain on MVTec AD,
and 12.8\% $F_1$ gain,
5.7\% NMI gain,
and 10.8\% ARI gain on MTD.
Code is available at \href{https://github.com/HUST-SLOW/AnomalyNCD}{https://github.com/HUST-SLOW/AnomalyNCD}.
\end{abstract}

\section{Introduction}

\renewcommand{\thefootnote}{}
\footnotetext{$*$ Contributed Equally. $\dagger$ Corresponding Author.}
\renewcommand{\thefootnote}{\arabic{footnote}}

\begin{figure*}[t]
\vspace{-0.5em}
\begin{center}
\centering
\includegraphics[width=1\textwidth]{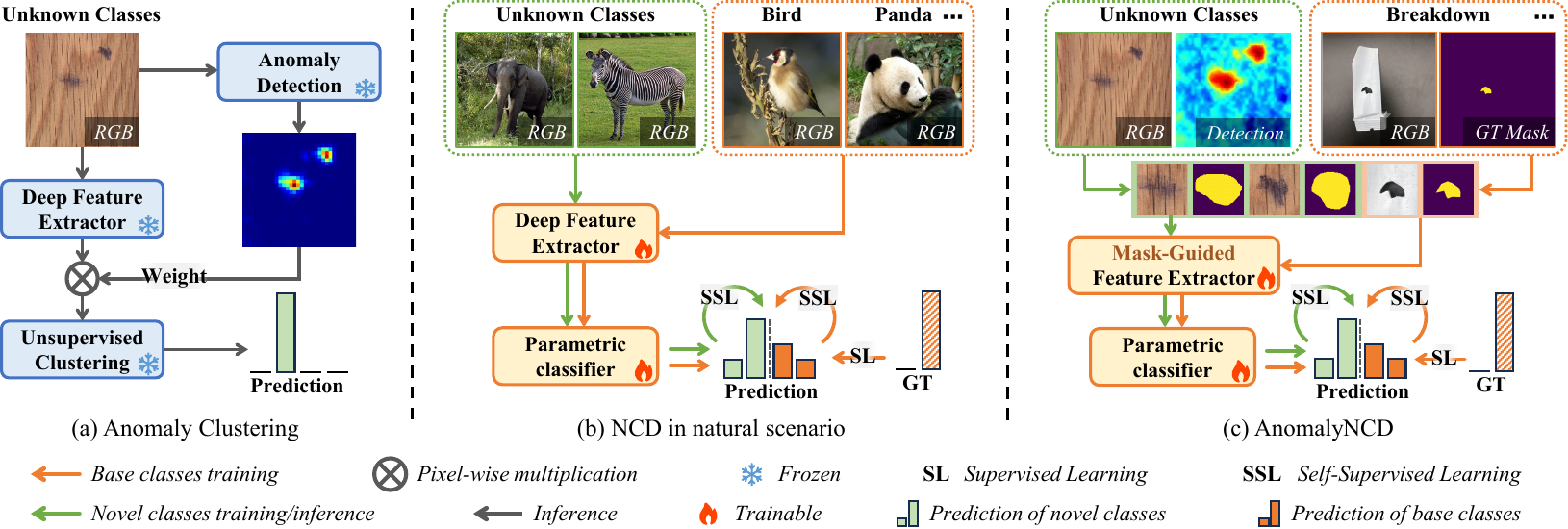}
\caption{\textbf{Comparison between solutions organizing anomalies into groups.}
(a) Anomaly clustering methods extract the features of the anomaly region and employ unsupervised clustering algorithms to cluster the anomalies.
(b) Vanilla NCD methods typically employ a trainable feature extractor and classifier. These components process object-centered images from both known and unknown classes.
(c) Our method aims to learn features of isolated anomaly regions using anomaly-centered sub-images and masks from MEBin.}
\vspace{-17pt}
\label{intro_2}
\end{center}
\vspace{-0.5em}
\end{figure*}

In recent years,
industrial anomaly detection \cite{CVPR2023winclip, tip2023ofcsAD, tip2021LSTDD, tip2021RFDD} has garnered remarkable performance.
It locates visual anomalies on industrial products but without recognizing fine-grained anomaly categories.
For downstream anomaly treatments,
it is necessary to recognize the anomaly classes,
such as fracture, ablation, etc.,
and even discover novel anomaly categories that are constantly emerging.

Several anomaly clustering approaches \cite{WACV2023AC,Arxiv2023uniformaly,CVPRW2024Blind} have been proposed to address the multi-class anomaly classification task, as illustrated in Fig.~\ref{intro_2} (a).
These approaches typically employ a two-step process: anomaly localization followed by feature extraction from the anomalous regions for clustering.
However, the effectiveness of these methods is limited when dealing with homogeneous anomaly patterns that exhibit differences in shape, appearance, and location,
primarily due to the lack of prior knowledge of anomaly types.
Therefore, our study aims to classify unseen anomalies by leveraging prior knowledge of seen anomalies, thereby overcoming the limitations of current methods.

Through detailed investigation, we find two primary obstacles in leveraging prior knowledge of known anomalies,
which prevent existing classification network to industrial anomalies. 
\ding{182} \textbf{Non-prominence Anomalies}:
Classification in natural scenes assumes that subjects are centered in the images so that networks easily extract their semantics,
while this assumption is not true in industrial scenarios.
\ding{183} \textbf{Low-semantics Anomalies}:
In contrast to natural objects,
industrial anomalies are typically less semantic,
making the network hard to focus on anomalies but prone to background.

In this paper, we overcome both obstructions and propose a multi-class anomaly classification network that aligns with the concept of novel class discovery (NCD), 
as illustrated in~\ref{intro_2} (c).
This network learns to classify isolated anomalies by focusing the model's attention on real anomaly regions.
Firstly, to address the non-prominence issue, 
we propose a main element binarization (MEBin) method.
MEBin isolates the primary anomaly regions by extracting them from anomaly detection (AD) results.
This binarization process alleviates the negative impact of errors within AD results on the subsequent multi-class anomaly learning,
enabling compatibility with various AD algorithms.
Secondly,
to tackle low-semantics anomalies, 
we propose mask-guided representation learning.
It leverages the anomaly masks generated by MEBin to direct the network's attention specifically to individual anomalous regions,
ensuring our network learns more discriminative features specific to anomalies and classifies each individual anomaly.
Additionally,
we use corrected pseudo labels in this learning method to prevent false-positive anomalies from training.
Finally,
to achieve more flexible classification in both region and image levels,
we propose a region merging strategy,
which determines the image category according to the classification of each region in the image.
Experimental results on the MVTec AD \cite{CVPR2019mvtec} and MTD datasets \cite{VC2020MTD} demonstrate the effectiveness of AnomalyNCD.

In summary, the major contributions of our work are:
\begin{itemize}

\item We propose the first method based on self-supervised manner in multi-class anomaly classification, named AnomalyNCD.
It can be seamlessly combined with different anomaly detection methods and classify unseen anomalies detected into homogeneous groups.

\item
We study the challenges of learning from anomalous regions in industrial scenarios,
motivating us to propose MEBin and mask-guided representation learning for our network to focus on real anomalous regions and learn discriminative features of isolated anomalies.

\item
AnomalyNCD surpasses existing anomaly clustering methods and NCD methods on the MVTec AD and MTD datasets,
providing a decent foundation for downstream anomaly treatments in industrial scenarios.

\end{itemize}

\section{Related Work}
\subsection{Image Clustering in Industrial Scenarios}
Directly clustering the images cannot perform well in multi-class anomaly classification.
To address this issue, two studies build the clustering pipeline specific for fine-grained anomaly classes.
Sohn \textit{et al.} \cite{WACV2023AC} proposed a simple but effective method to aggregate the patch features by anomaly map weighting.
Lee  \textit{et al.} \cite{Arxiv2023uniformaly} only aggregates the patch features with the higher anomaly score to determine the classes.
However, both methods rely on frozen models that cannot learn features specific to anomalies.
Therefore, we adopt self-supervised learning to discover novel classes from unlabeled anomalies.

\begin{figure*}[!t]
\vspace{-0.5em}
\begin{center}
\includegraphics[width=1\textwidth]{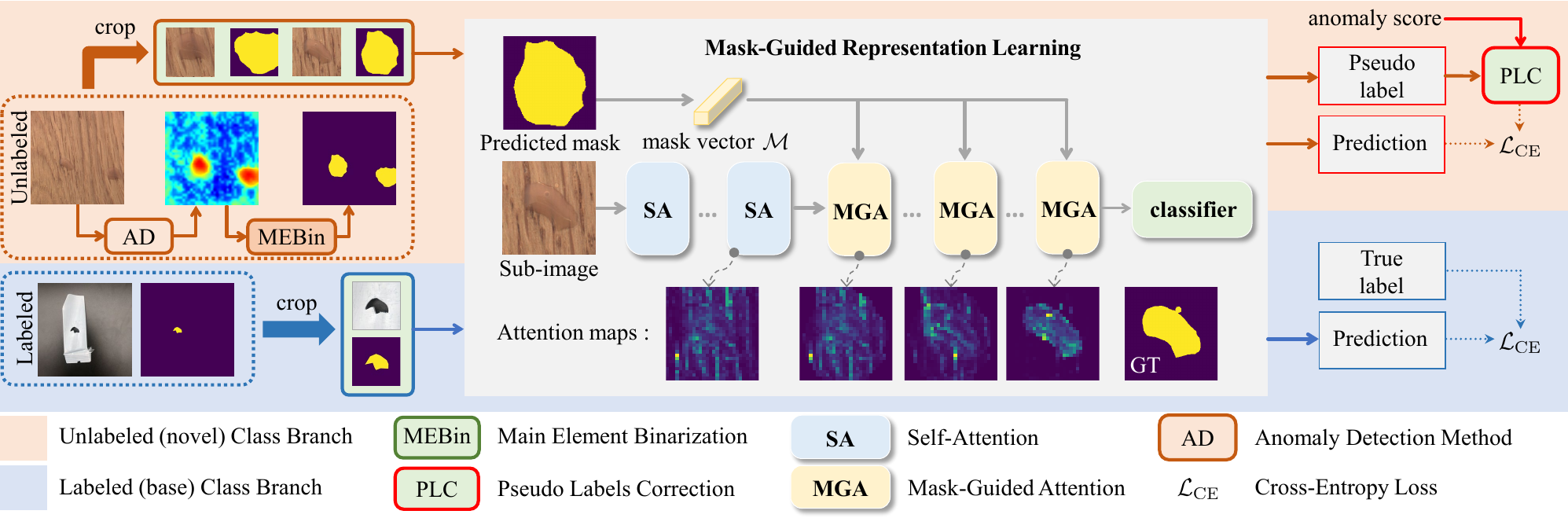}
\vspace{-1.4em}
\caption{
\textbf{Illustration of the Training Process for AnomalyNCD}.
First, we apply main element binarization (MEBin) to segment anomaly masks from detection results and generate anomaly-centered sub-images.
Second, we introduce mask-guided representation learning to learn discriminative features of anomalies, classifying sub-images into various categories.
}
\vspace{-9pt}
\label{pipeline}
\end{center}
\vspace{-0.5em}
\end{figure*}

\subsection{Novel Class Discovery}
Novel Class Discovery setting was first formalized in \cite{ICCV2019DTC}.
The early methods follow a two-stage pipeline \cite{ICCV2019DTC,ICLR2018transfer,ICLR2019multi}.
The first stage is to learn representation from the base set to obtain prior knowledge for classification,
which serves as the basis for the second one to train the model on the novel set.
Recent methods are one-stage \cite{TPAMI2021autonovel, CVPR2021openmix, NIPS2023uGCD}.
By handling the base set and novel set simultaneously,
one-stage methods extract better representations with less bias towards the base classes.
NCD has been applied in several fields,
such as medical image classification \cite{CVPR2024SeeingUnseen, AAAI2024Xray}, semantic segmentation \cite{CVPR2022ncdss} and point cloud segmentation \cite{CVPR2023pcs, ECCV2025pcs}.
In this paper,
we introduce the NCD architecture into multi-class industrial anomaly classification for the first time.

\subsection{Binarization Approach}
Binarization is a crucial preprocessing step for various vision tasks,
e.g., edge detection \cite{CVPR2015DeepEdge, VI2023occlusion},
segmentation \cite{TITS2022fastroadseg,ICRA2022fast,TITS2024ufnet},
document image processing \cite{TIP2019GiB, TIP2012bin, tip2013PEM}, 
medical image processing \cite{TMI2003binary, SIP2012MRIbinary},
object detection \cite{tip2020tiny,ICRA2019tiny,IJCV2024indoor}, etc.
Otsu \cite{TSMC1979otsu} determines the optimal threshold to separate the foreground and background by maximizing the between-class variance of the pixel intensities.
DiffuMask \cite{ICCV2023diffumask} utilizes cross-attention maps and adaptive thresholding to generate binary maps.
However, these methods usually obtain fragmented false-positive regions when segmenting the anomalies,
which harms the region-level anomaly class prediction.
To alleviate this,
we propose the main element binarization strategy to extract only the primary anomaly region,
thereby reducing the impact of false positives on representation learning.

\section{Method}
\label{sec:method}
The proposed AnomalyNCD aims to automatically discover and recognize the visual categories of industrial anomalies.
Fig.~\ref{pipeline} illustrates its overall framework.
To extract main anomaly regions from detection results,
we outline the main element binarization method (Sec.~\ref{sec:binarization}).
Guided by these anomaly regions, we then apply mask-guided representation learning (Sec.~\ref{sec:mask_guide}) to extract and classify discriminative features of anomalies within these regions.
Finally, a region merging strategy (Sec.~\ref{sec:merge}) ensures robust image-level classification based on all anomaly regions.

\subsection{Problem Definition} 
\label{sec:Preliminaries}
Given a set of unlabelled images $\mathcal{D}^\mathbf{u}=\{ I_i^\mathbf{u} | i\in[1,N^\mathbf{u}]\}$,
the goal of AnomalyNCD is to assign them into $\mathcal{C}^\mathbf{u}$ classes (named ``\textbf{Novel}'' previously),
where $\mathcal{C}^\mathbf{u}$ is known as a priori,
referring to vanilla NCD \cite{ICCV2021UNO, CVPR2022GCD, ICCV2023Class-relation}.
To provide a better foundation for learning novel classes,
we follow previous NCD tasks to assume a set of labeled anomaly images $\mathcal{D}^\mathbf{l}=\{(I_i^\mathbf{l}, y_i^\mathbf{l}, \mathbf{M}_i^\mathbf{l})|i\in[1,N^\mathbf{l}]\}$ with $\mathcal{C}^\mathbf{l}$ classes available (named ``\textbf{Base}'' previously),
where $y_{i}^\mathbf{l} \in \mathbb{R}^{1 \times (\mathcal{C}^\mathbf{l}+\mathcal{C}^\mathbf{u})}$ is the one-hot class label,
and $\mathbf{M}_{i}^\mathbf{l}$ is the ground truth anomaly mask of image $I_i^\mathbf{l} \in \mathcal{D}^\mathbf{l}$.
$\mathcal{D}^\mathbf{l}$ supplies the auxiliary knowledge of the industrial anomalies for grouping the images in $\mathcal{D}^\mathbf{u}$.
To further generalize our study, we also perform experiments with different $\mathcal{D}^\mathbf{l}$ and without $\mathcal{D}^\mathbf{l}$ in the Appendix F.

\subsection{Main Element Binarization}
\label{sec:binarization}
To provide a clear anomaly indication of unlabeled images,
we use an anomaly detection method on $I^\mathbf{u}_i$ to generate the anomaly probability map, denoted as $A_i\!\in\![0,1]^{H \times W}$.
However, $A_i$ inevitably contains false positives (over-detections) and false negatives (missed detections).
These errors can negatively affect multi-class anomaly classification.
To address this, we propose a Main Element Binarization (MEBin) approach.
It extracts the principal anomaly structures (Main Element) from $A_i$ by focusing on stable regions across small threshold variations.
Overall,
the MEBin consists of three steps:
\ding{182} Predefine a range $[\mathbf{s}_{\min},\mathbf{s}_{\max}]$ for threshold exploration;
\ding{183} Binarize $A_i$ across all thresholds within this range;
\ding{184} Identify the main elements in $A_i$ based on the changes of segmented regions under different binarization results.

\textbf{The first step:}
To capture all potential anomalies across the set $\{A_i|i\!\!\in\!\![1,N^\mathbf{u}]\}$,
the upper limit $\mathbf{s}_{\max}$ is set to 1.
Conversely, the lower limit, $\mathbf{s}_{\min}$,
is determined by the minimum anomaly score\footnote{Anomaly score: Maximum value in an anomaly probability map.} found across the set $\{A_i|i\!\!\in\!\![1,N^\mathbf{u}]\}$,
represented as $\mathbf{s}_{\min} \!=\! \min(s_1, s_2, ..., s_{N^\mathbf{u}})$, where $s_i \!=\! \max(A_i)$.
Such a lower limit helps minimize the likelihood of normal image pixels being segmented.

\begin{figure}[!t]
\vspace{-0.5em}
\centering
\includegraphics[width=0.47\textwidth]{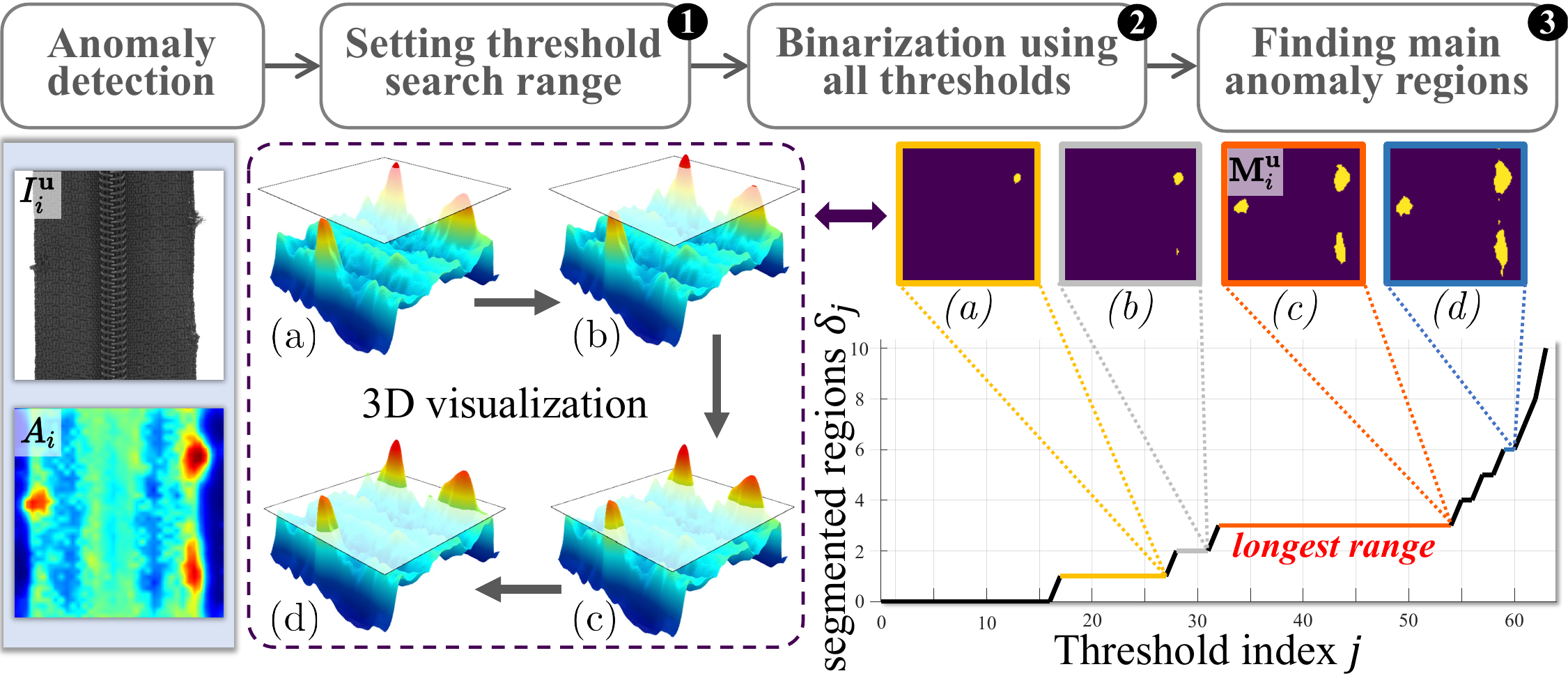}
\caption{
\textbf{Pipeline of the Main Element Binarization.} In 3D visualization, we show the changes of segmented regions under different thresholds. The 3D map $\textup{(a)}$ corresponds to the 2D binary segmentation mask $\emph{(a)}$.
}
\label{fig:binarization}
\vspace{-0.8em}
\end{figure}

\textbf{The second step:} 
To analyze the segmented regions change with thresholds, 
we uniformly sample $\mathcal{T}$ thresholds $\{\epsilon_j|j\in[1,\mathcal{T}]\}$ in $[\mathbf{s}_{\min},\mathbf{s}_{\max}]$.
For each threshold $\epsilon_j$,
the anomaly map $A_i$ is binarized into a mask $M_i^{j}$:
$M_i^{j} = \mathds{1}[A_i > \epsilon_j]$,
where $\mathds{1}[.]$ is the indicator function.
To reduce the meaningless and fragmented segmentation,
we further apply an erosion operation to each mask $M_i^{j}$.

\textbf{The third step:} We identify main anomaly regions through two operations.
First,
let $\delta_i^j$ denote the number of individual regions segmented in $M_i^j$,
each region is an independent connected component.
We observe that the amount of the main regions in the anomaly map $A_i$ can be characterized by the non-zero value that appears most frequently in $\{\delta_i^j|j\!\!\in\!\![1,\mathcal{T}]\}$,
denoted as $\bar{\delta}_i$.
Second, to segment the main regions,
we take the longest continuous threshold range where only $\bar{\delta}_i$ regions are segmented.
The main anomaly regions remain stable within this threshold range, as shown in Fig.~\ref{fig:binarization}. 

\begin{itemize}
\item If this range is shorter than $\tau$, 
there is no anomaly region to be segmented.
\item 
Conversely, if this range exceeds $\tau$,
it confirms stable anomaly segmentation within this threshold range.
In this case,
we select the minimum threshold from the identified range to segment anomalies as completely as possible.
\end{itemize}
Finally, we use the selected threshold to segment $A_i$ into the binary mask, denoted as $\mathbf{M}_i^\mathbf{u}$.
In this way,
the unlabeled images set $\mathcal{D}^{\mathbf{u}}$ are extended with the binary masks: $\{(I_i^\mathbf{u},\mathbf{M}_i^\mathbf{u})|i\in[1,N^\mathbf{u}]\}$.

\noindent
\textbf{Anomaly-Centered Sub-Image Cropping.}
To focus the model on anomalous regions, 
we crop each region in $\mathbf{M}_i^\mathbf{u}$ individually into anomaly-centered sub-images.
Specifically,
for each region in $\mathbf{M}_i^\mathbf{u}$,
we use the minimum bounding square to crop the sub-image from $I_i^\mathbf{u}$ and its mask from $\mathbf{M}_i^\mathbf{u}$.
Then the sub-images of $I_i^\mathbf{u}$ can be denoted as $\{x_{i,1}^\mathbf{u},x_{i,2}^\mathbf{u},\ldots\}$,
and the corresponding masks cropped from $\mathbf{M}_i^\mathbf{u}$ are denoted as $\{m_{i,1}^\mathbf{u},m_{i,2}^\mathbf{u},\ldots\}$.
To align labeled and unlabeled inputs,
we use the same cropping operation to extract sub-images and masks for the labeled data set $\mathcal{D}^\mathbf{l}$,
denoted as $\{(x_{i,1}^\mathbf{l},m_{i,1}^\mathbf{l}), (x_{i,2}^\mathbf{l},m_{i,2}^\mathbf{l}), \ldots |i\in[1,N^\mathbf{l}]\}$.
These are then merged with the unlabeled data in the same structures,
represented as $\{(x_{i,k},m_{i,k},y_{i,k})|i\!\in\![1,N^\mathbf{l}\!+\!N^\mathbf{u}],k\!\in\![1,\bar{\delta}_i]\}$,
where $\bar{\delta}_i$ indicates the number of sub-images per image,
and $y_{i,k}$ is the one-hot class label for each labeled sub-image,
matching the label of the $i^\text{th}$ image if it belongs to $\mathcal{D}^\mathbf{l}$.
Implementation details of the cropping process are supplied in the Appendix C.

\subsection{Mask-Guided Representation Learning}
\label{sec:mask_guide}

\begin{figure}[!t]
\vspace{-0.5em}
\begin{center}
\includegraphics[width=0.45\textwidth]{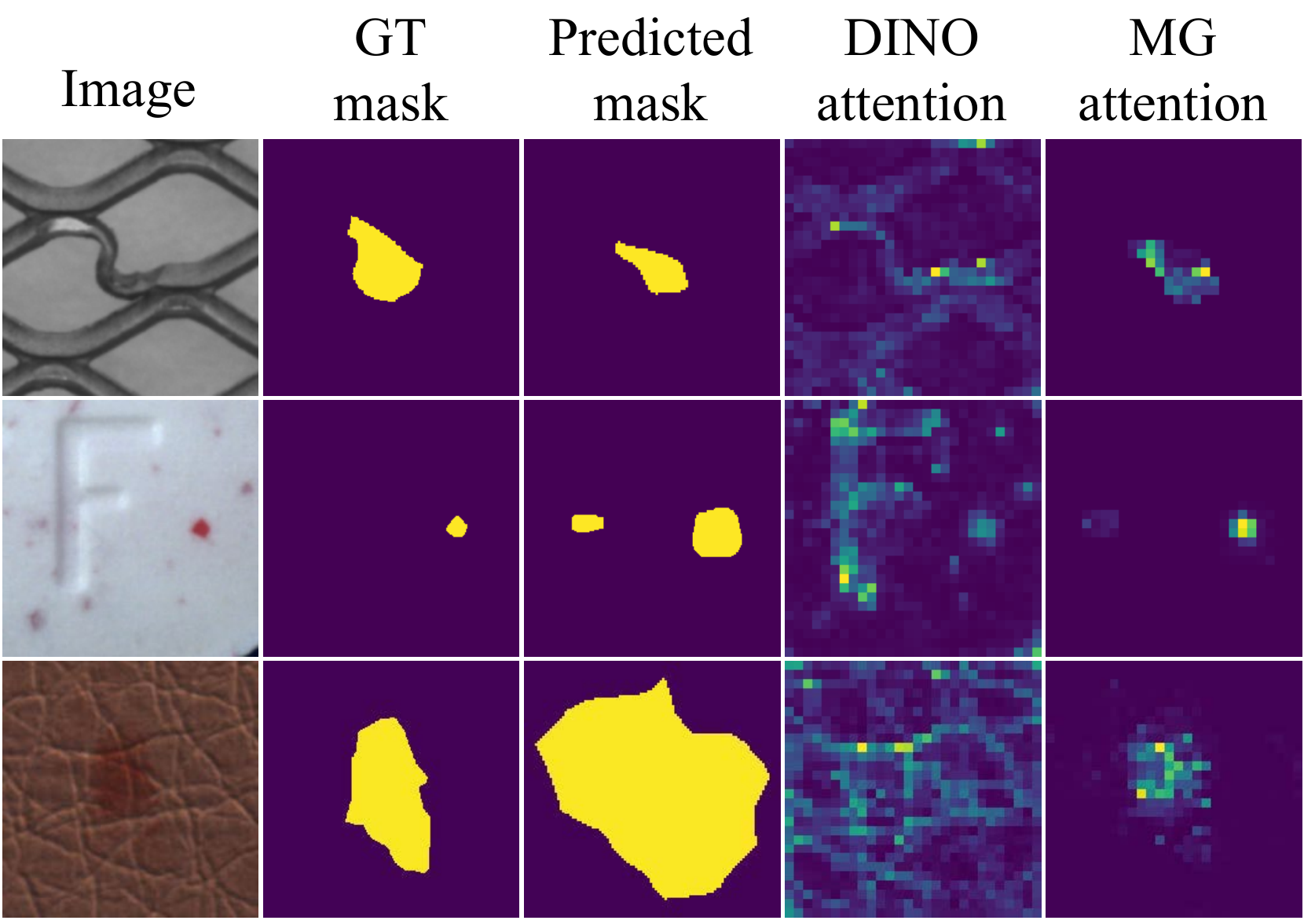}
\caption{
\textbf{Visualization of the self-attention of the $\texttt{[CLS]}$ token on the last layer's heads.}
DINO attention refers to the $\texttt{[CLS]}$ token extracted from a DINO pre-trained ViT that mainly focuses on a foreground object.
AnomalyNCD uses a mask to direct the $\texttt{[CLS]}$ token's attention to the anomalous regions.}
\vspace{-13pt}
\label{attention_vis}
\end{center}
\vspace{-0.5em}
\end{figure}

\subsubsection{Mask-Guided Vision Transformer (MGViT)}
\label{sec:mgvit}
MGViT is built on vision transformer (ViT).
Thus we review its structure.

\noindent
\textbf{Vision transformer} \cite{ICLR2021ViT} processes an input image by dividing it into \texttt{N} fixed-size patches.
Each patch is treated as a ``token'' and is flattened into an embedding vector of length $D$.
ViT has $L$ sequential transformer layers,
each containing self-attention layers and feed-forward neural networks.
In addition, a special class token  $\texttt{[CLS]}$,
is propagated from the first layer to the final layer where it serves as the basis for classification.
In the self-attention of the $l^\text{th}$ layer,
the input patches undergo linear projections to form the queries, keys, and values $\mathbf{Q}_{l-1}, \mathbf{K}_{l-1}, \!\mathbf{V}_{l-1} \!\! \in \! \mathbb{R}^{(\texttt{N}+1)\times D}$.
Here, ViT captures the relationships between the image patches by self-attention mechanism $Attn$ as follows:
\begin{equation}
\!\!Attn\!=\!\text{softmax}\big(\text{concat}(\mathbf{Q}^{\text{cls}}_{l-1}\mathbf{K}^{\top}_{l-1}, \mathbf{Q}^{\text{patch}}_{l-1}\mathbf{K}^{\top}_{l-1})\big)\mathbf{V}_{l-1}
\end{equation}
where we divide $\mathbf{Q}_{l-1}$ into $\mathbf{Q}^{\text{cls}}_{l-1}\!\!\in \!\! \mathbb{R}^{1\times D}$ and $\mathbf{Q}^{\text{patch}}_{l-1}\!\!\in \!\! \mathbb{R}^{\texttt{N}\times D}$ to represent class token and patch tokens separately.

\noindent
\textbf{Insufficient Focus on Anomalies.}
The ViT excels in semantic representation but fails to detect weak semantic anomalies.
Fig.~\ref{attention_vis} demonstrates this using the self-attention map of the $\texttt{[CLS]}$ token in the last layer of a DINO-pretrained ViT \cite{iccv2021dino}.
In the $1^\text{st}$ row’s example,
the model highlights the salient object (metal grid) in an image, rather than the local anomaly.
This observation means that the network over-focuses on learning background.

\noindent
\textbf{Mask-Guided Attention.}
To address the issue above,
we propose mask-guided attention to direct ViT's focus on the anomalous region.
Specifically,
the input sub-image $x_{i,k}$ is fed into the ViT, and we assume the token number is $\texttt{N}$.
At the same time,
its mask $m_{i,k}$ is resized to $\sqrt{\texttt{N}} \!\times\! \sqrt{\texttt{N}}$ by average pooling and then flattened into a vector $\mathcal{M} \in \mathbb{R}^{\texttt{N} \times 1}$.
Subsequently,
we insert the constant $1$ at the beginning of $\mathcal{M}$ to align its size with the tokens.
Next,
we use the mask vector in the self-attention mechanism for re-directing focus,
which has three possible design choices.
\begin{enumerate}
\item Adding the mask $\mathcal{M}$ on both class and patch tokens.
\item Adding the mask $\mathcal{M}$ on the patch tokens.
\item Adding the mask $\mathcal{M}$ on the class token.
\end{enumerate}
The $1^\text{st}$ and $2^\text{nd}$ designs suppress the rich contextual features in patch tokens that are crucial for classification,
and our network only focuses on classifying anomalies rather than segmenting.
Thus, the $3^\text{rd}$ one is recommended as the most suitable design,
and the experiment also verifies its effectiveness.
Such a design can be formulated as follows:
\begin{equation}
\!\!Attn\!=\!\text{softmax}(\!\text{concat}(\!\mathbf{Q}^{\text{cls}}_{l-1}\mathbf{K}^{\top}_{l-1}\!+\!\overline{\mathcal{M}}\,, \mathbf{Q}^{\text{patch}}_{l-1}\mathbf{K}^{\top}_{l-1}\!)\!)\mathbf{V}_{\!l-1}
\end{equation}
where $\overline{\mathcal{M}}$ denotes the converted mask vector that is compatible with the attention mechanism \cite{CVPR2022mask2former}:
\begin{align}
\overline{\mathcal{M}}(i) = 
\begin{cases}
0, & \quad \text{if~} \mathcal{M}(i)>0.5\\
-\infty, & \quad \text{otherwise}
\end{cases}
\end{align}
Note that,
in the ViT,
the self-attention in the last $L_m$-layer is replaced with the mask-guided attention to form MGViT.
Fig.~\ref{attention_vis} reveals that the $\texttt{[CLS]}$ token focuses on anomalous regions by using our mask-guided attention.
Even though the predicted masks do not align with the ground truth accurately,
the network still focuses on the anomaly roughly.

\subsubsection{Model Training}
\label{sec:training}
Given a sub-image pair $(x_{i,k},m_{i,k})$, we first apply random augmentation to
generate two views $(\tilde{x}_{i,k},\tilde{m}_{i,k})$ and $(\hat{x}_{i,k},\hat{m}_{i,k})$.
Following DINO \cite{iccv2021dino}, we feed these two views into the teacher and student networks respectively,
which have shared MGViT and classification heads with different softmax temperatures.
For labeled sub-images, the ground truth labels are converted into one-hot vectors to serve as supervision targets for the student network.
For unlabeled sub-images $\tilde{x}_{i,k},\hat{x}_{i,k}$, we generate pseudo labels $\hat{q}_{i,k},\tilde{q}_{i,k} \in \mathbb{R}^{1 \times (\mathcal{C}^\mathbf{l}+\mathcal{C}^\mathbf{u})}$ by the ``teacher'' one, which employs a sharp temperature $\tau_t$ to produce confident predictions.
The entries of labeled classes ($\mathcal{C}^\mathbf{l}$) in the pseudo labels are set to $0$.
The ``student'' one, using a smooth temperature $\tau_s$, is trained to align its probability predictions $\hat{p}_{i,k}$, $\tilde{p}_{i,k} \in \mathbb{R}^{1 \times (\mathcal{C}^\mathbf{l}+\mathcal{C}^\mathbf{u})}$ with either ground truth labels (for labeled data) or pseudo labels (for unlabeled data).
This dual supervision ensures effective learning from both labeled and unlabeled data.

Our training objective follows existing category discovery methods \cite{ICCV2023SimGCD,CVPR2022GCD} and includes classification objective and representation learning objective.
The classification objective contains ground truth label supervision $\mathcal{L}_{\text{cls}}^\mathbf{l}$ on labeled sub-images, and pseudo label supervision $\mathcal{L}_{\text{cls}}^\mathbf{u}$ on unlabeled sub-images.
The representation learning objective contains self-supervised contrastive learning $\mathcal{L}_{\text{rep}}$ \cite{ICML2020SimCLR} on both labeled sub-images and unlabeled sub-images, and supervised contrastive learning $\mathcal{L}_{\text{rep}}^\mathbf{l}$ \cite{NIPS2020supervised} on only labeled sub-images.
Finally, we adopt a mean-entropy maximization regularizer $\mathcal{L}_{\text{reg}}^\mathbf{u}$ \cite{NIPS2020entropy1} for unlabeled sub-images.
More details about the training procedure are provided in Appendix A.

The overall loss of training objective is written as:
\begin{equation}
\mathcal{L} = \lambda(\mathcal{L}_{\text{rep}}^\mathbf{l}+\mathcal{L}_{\text{cls}}^\mathbf{l}) +(1-\lambda)(\mathcal{L}_{\text {rep}} + \mathcal{L}_{\text{cls}}^\mathbf{u}+\mu \mathcal{L}_{\text{reg}}^\mathbf{u})
\end{equation}
where $\lambda$ is a hyperparameter that balances the supervised loss and the self-supervised loss.
$\mu$ is the coefficient of the regularization.

\noindent
\textbf{Pseudo Labels Correction.}
\label{sec:pseudo_labels}
The error in pseudo labels leads to the false learning objective for the classification model.
To mitigate these errors, 
especially the false detection of anomalous regions, 
we propose to use anomaly scores (described in Sec~\ref{sec:binarization}) to correct pseudo labels.

Let $s_{i,k}$ denote the anomaly score for $x_{i,k}$,
we weightedly judge whether the view $\hat{x}_{i,k}$ should be labeled as normal or assigned the generated pseudo labels.
\begin{equation}
\!\!\hat{q}_{i,k} \!\leftarrow\! w_{i,k}\mathbf{e}+(1-w_{i,k})\hat{q}_{i,k}
;\,\,\,w_{i,k} \!=\! \mathop{\max}(0.5-s_{i,k},0)
\end{equation}
where $w_{i,k}$ is the weight for normal.
$\mathbf{e}\!\in\!\mathbb{R}^{1 \times (\mathcal{C}^\mathbf{l}+\mathcal{C}^\mathbf{u})}$ denotes a unit vector whose ($\mathcal{C}^\mathbf{l}+1$)-th value is $1$ and all other values are $0$,
in which we use the first class in the $\mathcal{C}^\mathbf{u}$ classes as normal.
If $s_{i,k}$ is smaller,
the more likely the sub-image $\hat{x}_{i,k}$ is normal,
and $\hat{q}_{i,k}$ is closer to $\mathbf{e}$.
Thus,
such correction forces the network to separate normal sub-images from the train set.
In the same way,
the pseudo label $\tilde{q}_{i,k}$ is updated.

\begin{figure}[t]
\begin{center}
\vspace{-0.5em}
\includegraphics[width=0.5\textwidth]{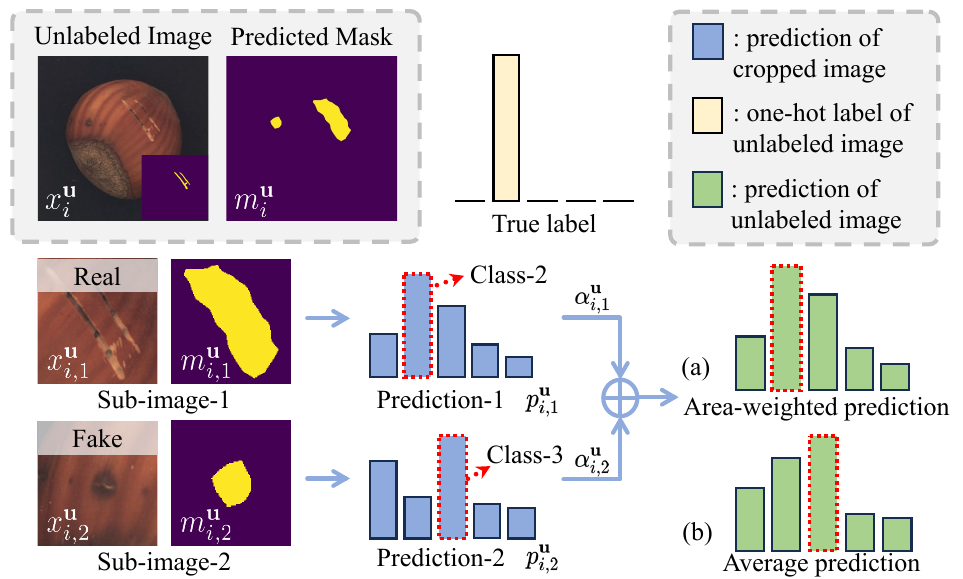}
\caption{
\textbf{Region merging strategy for image-level classification.}
Using average prediction, the output of normal cropped images (Sub-image-2) leads to final misclassification, while area-weighted prediction can reduce the negative effect of normal cropped images on the final result.}
\vspace{-8pt}
\label{merge_vis}
\end{center}
\vspace{-1.5em}
\end{figure}

\subsection{Region Merging for Image Classification}
\label{sec:merge}
For flexible classification in both sub-image and image levels,
we aim to determine the image's anomaly class according to its sub-image classification.
Intuitively,
a naive method is to average the predictions of all sub-images within an image.
However,
false classification of the over-detected region may mislead the image class.
As shown in Fig.~\ref{merge_vis},
\textit{Sub-image-1} and \textit{Sub-image-2} are classified to \textit{Class-2} and \textit{Class-3}, respectively,
where the wrong prediction of \textit{Sub-image-2} misleads the image class in Fig.~\ref{merge_vis} (b).

To address this issue,
we propose a region merging approach to robustly classify images.
We found that misclassification is mostly over-detected regions that are much smaller than real anomalies.
Therefore, we define an area-related weight $\alpha_{i,k}^\mathbf{u}$ for each sub-image $x_{i,k}^\mathbf{u}$,
which is used in determining the image category:
\begin{equation}
\alpha_{i,k}^\mathbf{u}=\frac{\exp( {\sqrt{\textbf{a}_{i,k}^\mathbf{u}}} / {\tau_\alpha})}
{\sum_{k=1}^{\bar{\delta}_i} \exp( {\sqrt{\textbf{a}_{i,k}^\mathbf{u}}} / {\tau_\alpha})}
\end{equation}
where $\textbf{a}_{i,k}^\mathbf{u}$ is the area of anomaly in $x_{i,k}^\mathbf{u}$,
and $\tau_\alpha$ is the temperature value.
Then,
for an image $I_i^\mathbf{u}$ composed by sub-images $\{x_{i,k}^\mathbf{u}|k\!\in\![1,\bar{\delta}_i]\}$,
the prediction logit is $p_i^\mathbf{u}=\sum\nolimits_{k=1}^{\bar{\delta}_i}{\alpha_{i,k}^\mathbf{u} p_{i,k}^\mathbf{u}}$,
where $p_{i,k}^\mathbf{u}$ is the probability prediction of $x_{i,k}^\mathbf{u}$ generated by the student network.
As shown in Fig.~\ref{merge_vis} (a), the prediction $p_i^\mathbf{u}$ depends more on $p_{i,1}^\mathbf{u}$, which has a larger weight $\alpha_{i,1}^\mathbf{u}$ and a correct prediction.

\begin{table*}[!t]
    \vspace{-1em}
  \centering
      \setlength\tabcolsep{4pt}
        \resizebox{1.0\linewidth}{!}{    \begin{tabular}{cccccccccccc}
    \toprule
          & \multirow{2}{*}{Datasets} & \multirow{2}{*}{Metric} & \multirow{2}{*}{\textbf{IIC\cite{ICCV2019IIC}}} &  \multirow{2}{*}{\textbf{GATCluster\cite{ECCV2020gatcluster}}} &  \multirow{2}{*}{\textbf{SCAN\cite{ECCV2020scan}}} &  \multirow{2}{*}{\textbf{UNO\cite{ICCV2021UNO}}} & \multirow{2}{*}{\textbf{GCD\cite{CVPR2022GCD}}} & \multirow{2}{*}{\textbf{SimGCD\cite{ICCV2023SimGCD}}} &  \multirow{2}{*}{\textbf{AMEND\cite{WACV2024amend}}} &  \textbf{AC\cite{WACV2023AC}} & \textbf{MuSc \cite{ICLR2024MuSc}} \\
          &  &  &  &  &  &  &   &   &   & \textbf{(Unsup.)} & \textbf{+AnomalyNCD} \\
    \midrule 
    & \multirow{4}{*}{MVTec AD \cite{CVPR2019mvtec}}  & NMI & 0.093 & 0.136 & 0.210 & 0.146  & 0.417 & 0.452 & 0.431 & \underline{0.525} & \textbf{0.613} \\
    &  & ARI & 0.020 & 0.053  & 0.103 & 0.052  & 0.302  &  0.346 & 0.333 & \underline{0.431} & \textbf{0.526} \\
    &  & $F_1$ & 0.285 & 0.264 & 0.335 & 0.342 & 0.553 & 0.569 & 0.542 & \underline{0.604} & \textbf{0.712} \\
    \midrule
    & \multirow{4}{*}{MTD \cite{VC2020MTD}}  & NMI & 0.064 & 0.028 & 0.041 & 0.034  & \underline{0.211} & 0.105 & 0.138 & 0.179 & \textbf{0.268} \\
    &  & ARI & 0.020 & 0.009 & 0.029 & 0.011 & 0.115 & 0.048 & 0.067 & \underline{0.120} & \textbf{0.228} \\
    &  & $F_1$ & 0.252 & 0.243 & 0.282 & 0.221 & \underline{0.381} & 0.293 & 0.324 & 0.346 & \textbf{0.509} \\
    \bottomrule
    \end{tabular}
    }
  \vspace{-0.5em}
  \caption{Quantitative results on the MVTec AD and MTD dataset. All the methods only use unlabeled images as input. The best-performing result is in bold, the second best result is underlined.}
  \label{tab:main_comp_1}
\end{table*}

\begin{table*}[!t]
  \centering
      \setlength\tabcolsep{4pt}
      \resizebox{1.0\linewidth}{!}{
        \begin{tabular}{cccccccccc}
    \toprule
          & \multirow{3}{*}{Datasets} & \multirow{3}{*}{Metric} & \textbf{AC\cite{WACV2023AC}} &  \textbf{UniFormaly\cite{Arxiv2023uniformaly}} & \textbf{PatchCore\cite{CVPR2022patchcore}} & \textbf{RD++\cite{cvpr2023rd}} & \textbf{EfficientAD\cite{wacv2024efficientad}} &\textbf{PNI\cite{iccv2023pni}} & \textbf{CPR\cite{TIP2024CPR}} \\
          &  &  & \textbf{(Semi-sup.)} & (0.953 / 0.837) & (0.938 / 0.729) & (0.950 / 0.741) & (0.917 / 0.731) & (0.942 / 0.516) & (0.964 / -) \\
          &  &  &  &  & \textbf{+AnomalyNCD} & \textbf{+AnomalyNCD} & \textbf{+AnomalyNCD} & \textbf{+AnomalyNCD} & \textbf{+AnomalyNCD} \\
    \midrule
    & \multirow{3}{*}{MVTec AD \cite{CVPR2019mvtec}}  & NMI & 0.608 & 0.547 & 0.670 & 0.631 & \colorbox[rgb]{0.9,0.9,0.9}{0.516} & \underline{0.675} & \textbf{0.736} \\
    &  & ARI & 0.489 & 0.433 & 0.601 & 0.542 & \colorbox[rgb]{0.9,0.9,0.9}{0.394} & \underline{0.609} & \textbf{0.674} \\
    &  & $F_1$ & 0.652 & 0.645 & \underline{0.769} & 0.721 & \colorbox[rgb]{0.9,0.9,0.9}{0.641} & \underline{0.769} & \textbf{0.805} \\
    
    \midrule
    & \multirow{3}{*}{MTD \cite{VC2020MTD}} & NMI & \underline{0.390} & \textbf{0.421} & 0.380 &  \colorbox[rgb]{0.9,0.9,0.9}{0.368} &  \colorbox[rgb]{0.9,0.9,0.9}{0.220} & \colorbox[rgb]{0.9,0.9,0.9}{0.181}  & - \\
    &  & ARI & 0.314 & 0.322 & \textbf{0.390} & \colorbox[rgb]{0.9,0.9,0.9}{\underline{0.361}} & \colorbox[rgb]{0.9,0.9,0.9}{0.188} & \colorbox[rgb]{0.9,0.9,0.9}{0.219}  & - \\
    &  & $F_1$ & 0.490 & \underline{0.609} & \textbf{0.617} & \colorbox[rgb]{0.9,0.9,0.9}{0.600} & \colorbox[rgb]{0.9,0.9,0.9}{0.467}  &  \colorbox[rgb]{0.9,0.9,0.9}{0.465} & - \\
    \bottomrule
    \end{tabular}
    }
    \vspace{-0.5em}
  \caption{Quantitative results on the MVTec AD and MTD dataset. All the methods use unlabeled images and labeled normal images as input. We evaluate the AUPRO metric across various anomaly detection approaches on two datasets, e.g., (0.953 / 0.837), the first one on the MVTec AD dataset and the second on the MTD dataset. For the metrics of multi-class anomaly classification, the best-performing result is in bold, the second best result is underlined. The unofficial results are marked in \colorbox[rgb]{0.9,0.9,0.9}{gray}.}
  \label{tab:main_comp_2}
  \vspace{-0.5em}
\end{table*}

\section{Experiments}

\subsection{Experimental setting}

\subsubsection{Datasets}
We conduct experiments on industrial datasets MVTec AD \cite{CVPR2019mvtec} and Magnetic Tile Defect (MTD) \cite{VC2020MTD}.
MVTec AD has industrial product images of 10 object categories and 5 texture categories.
Each product category has at least two anomaly classes.
Following \cite{WACV2023AC} and \cite{Arxiv2023uniformaly},
we remove the combined anomaly classes for a fair comparison.
Notice that our method can also handle the combined anomaly class, as described in the Appendix G.
MTD dataset has 952 normal images and 392 abnormal images, the abnormal images are divided into five anomaly classes.
We follow \cite{WACV2023AC} that uses 80\% of normal images as the reference images and the rest as the test images.
For both datasets, we use the single-blade sub-dataset in Aero-engine Blade Anomaly Detection Dataset (AeBAD-S) \cite{AeBAD} with the normal class removed, as our default labeled image set $\mathcal{D}^\mathbf{l}$.

\subsubsection{Implementation details}
We use ViT-B/8 \cite{ICLR2021ViT} pre-trained with DINO \cite{iccv2021dino} as our feature extractor.
The self-attentions in the last 9 layers are replaced with our mask-guided attention. 
During training, all the layers of ViT are fixed except the last layer.
For anomaly detection methods preceding AnomalyNCD, we choose zero-shot method MuSc \cite{ICLR2024MuSc}, one-class methods PatchCore \cite{CVPR2022patchcore}, EfficientAD \cite{wacv2024efficientad}, RD++ \cite{cvpr2023rd}, PNI \cite{iccv2023pni} and CPR \cite{TIP2024CPR}.
More details are in the Appendix B.

\subsubsection{Competing methods}
We compare our method with two state-of-the-art industrial anomaly clustering methods,
UniFormaly \cite{Arxiv2023uniformaly} and Anomaly Clustering \cite{WACV2023AC} that has two settings:
an unsupervised one (Unsup.) and a semi-supervised one (Semi-sup.).
The former setting only inputs unlabeled images for clustering,
while the latter one uses labeled normal images, same as the input of one-class anomaly detection.
We also employ three deep clustering methods,
IIC \cite{ICCV2019IIC}, GATCluster \cite{ECCV2020gatcluster}, and SCAN \cite{ECCV2020scan},
which cluster unlabeled images directly.
In addition,
four NCD methods in the nature scene are considered for comparison,
UNO \cite{ICCV2021UNO}, GCD \cite{CVPR2022GCD}, SimGCD \cite{ICCV2023SimGCD} and AMEND \cite{WACV2024amend}.
As in our method, the AeBAD-S dataset is used as labeled abnormal images.

\subsubsection{Evaluation metrics}
We report three widely used metrics for the evaluation of clustering:
the $F_1$ score, the normalized mutual information (NMI) \cite{NMI}, and the adjusted rand index (ARI) \cite{ARI}. 
We use the Hungarian algorithm \cite{hungarian} to match the predicted clusters to the ground truth labels.
For methods using anomaly maps,
we measure the segmentation per-region overlap (AUPRO) with 30\% FPR \cite{CVPR2019mvtec} to explore the impact of anomaly detection performance on the multi-class classification.
All metrics are calculated by the official code.

\begin{table}[!t]
    \centering
    \vspace{-8pt}
    \setlength{\tabcolsep}{2.3mm}
    \resizebox{1.0\linewidth}{!}{
    \begin{tabular}{l|cc|ccc}
        \toprule
        & Avg FPR $\downarrow$ & Avg FNR $\downarrow$ & NMI $\uparrow$ & ARI $\uparrow$ & $F_1$ $\uparrow$ \\ 
        \midrule 
        $\epsilon=0.1$ & 0.572 & 0.617 & 0.554 & 0.482 & 0.650  \\ 
        $\epsilon=0.3$ & 0.678 & 0.742 & 0.499 & 0.404 & 0.587  \\ 
        $\epsilon=0.5$ & 0.484 & 0.165 & 0.567 & 0.458 & 0.640  \\
        $\epsilon=0.7$ & 0.269 & 0.247 & 0.495 & 0.395 & 0.623   \\ 
        $\epsilon=0.9$ & 0.544 & 0.593 & 0.077 & 0.013 & 0.337   \\ 
        Otsu \cite{TSMC1979otsu} & 0.676 & 0.525 & 0.382 & 0.268 & 0.499   \\ 
        Ours & \textbf{0.153} & \textbf{0.035} & \textbf{0.613} & \textbf{0.526} & \textbf{0.712}   \\ 
        \bottomrule
    \end{tabular}
    }
    \vspace{-0.5em}
    \caption{Ablation study of different binarization approaches on the MVTec AD dataset. The best-performing result is in bold.}
    \label{tab:ablation_binarization}
    \vspace{-0.5em}
\end{table}

\subsection{Quantitative results}
We report the results of the multi-class anomaly classification on the MVTec AD and MTD datasets in Table~\ref{tab:main_comp_1} and~\ref{tab:main_comp_2}.
Notably, AnomalyNCD is compatible with all the anomaly detection methods to enable subsequent multi-class anomaly classification.

All methods in Table~\ref{tab:main_comp_1} only use unlabeled images as input.
Compared to Anomaly Clustering \cite{WACV2023AC},
AnomalyNCD combined with zero-shot anomaly detection method MuSc \cite{ICLR2024MuSc},
achieves 8.8\% gains on NMI, 9.5\% gains on ARI, and 10.8\% $F_1$ gains,
proving that AnomalyNCD extracts more discriminative features than Anomaly Clustering through contrastive learning in anomalous regions.
AnomalyNCD also outperforms state-of-the-art NCD methods,
achieving 16.1\% gains on NMI, 18.0\% gains on ARI, and 14.3\% $F_1$ gains.
The performance advantage of AnomalyNCD is due to focusing on anomalies inside images.

In Table~\ref{tab:main_comp_2},
all methods utilize both unlabeled images and labeled normal images from the same product.
We integrate AnomalyNCD with various one-class AD methods.
Utilizing CPR, which has the highest AUPRO,
achieves the best performance on the MVTec AD dataset:
NMI increases by 10.7\%, ARI by 14.9\%, and $F_1$ score by 9.6\%,
surpassing other methods.
On the MTD dataset,
although the segmentation AUPRO of all AD methods is lower than that of Uniformaly,
PatchCore+AnomalyNCD still achieves improvements of 6.8\% in ARI and 0.8\% in $F_1$ score.

\begin{figure}[!t]
\vspace{-0.5em}
\begin{center}  
\includegraphics[width=0.475\textwidth]{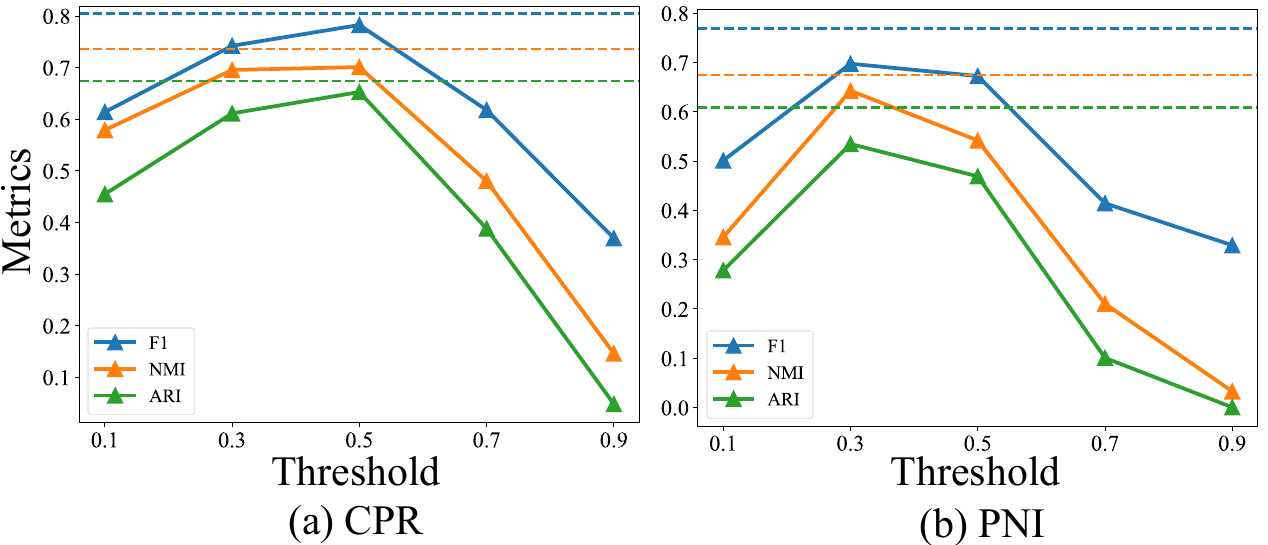}
\vspace{-2em}
\caption{
\textbf{The results of various anomaly detection (AD) methods under multiple fixed thresholds.}
The horizontal dashed line represents the results of MEBin. 
Notably, the optimal fixed thresholds differ across AD methods, 
while our MEBin consistently outperforms the fixed threshold results.
}
\vspace{-0.5em}
\label{table_vis}
\end{center}
\end{figure}

\begin{figure}[!t]
\vspace{-0.5em}
\begin{center}
\includegraphics[width=0.48\textwidth]{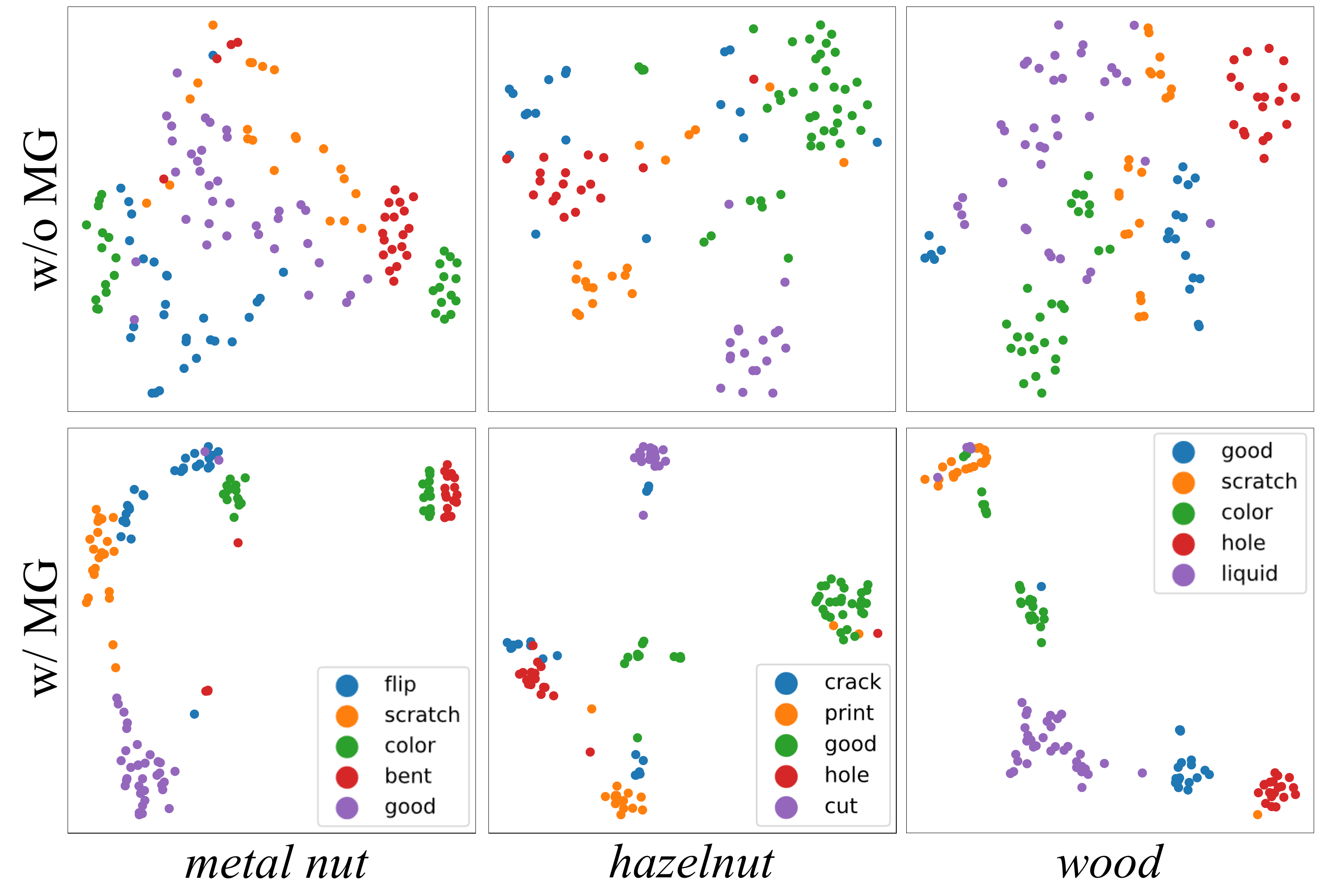}
\vspace{-1.5em}
\caption{
\textbf{T-SNE visualization of sub-images on the MVTec AD dataset.} We choose \emph{metal nut}, \emph{hazelnut} and \emph{wood} as examples. The different colors of dots represent their anomaly classes.
}
\vspace{-8pt}
\label{fig:tsne_vis}
\end{center}
\vspace{-1em}
\end{figure}

\subsection{Ablation Study}

\subsubsection{Effectiveness of main element binarization}

Table~\ref{tab:ablation_binarization} compares MEBin's performance with the Otsu method \cite{TSMC1979otsu} and fixed thresholds.
Since the missed detection is more critical than over-detection in industrial scenarios,
we define an anomaly as detected if the intersection over union (IoU) between the actual and predicted bounding boxes exceeds 0.1.
Then we calculate the FPR and FNR for these binarization methods.
MEBin demonstrates lower FPR and FNR than fixed thresholds and Otsu,
leading to better multi-class classification results.
Compared to the fixed thresholds, MEBin adaptively determines an optimal threshold for each image adaptation and achieves a 7.2\% $F_1$ improvement.
The Otsu tends to over-detect, particularly in normal images, making it less effective for this task.

In Fig.~\ref{table_vis},
we further apply MEBin on different one-class AD methods, which exhibit varying optimal fixed thresholds.
For example, the optimal threshold of CPR \cite{TIP2024CPR} is near 0.5, whereas for PNI \cite{iccv2023pni} it is approximately 0.3.
MEBin's adaptive threshold selection for each image outperforms all fixed thresholds for the whole image set.

\subsubsection{Discussion of the mask-guided attention}
The mask-guided attention (MGA) directs ViT to focus on tokens within the mask (anomalous region).
We show the t-SNE visualization in Fig.~\ref{fig:tsne_vis}.
With MGA in the $2^{nd}$ row, image features have larger inter-class distances (e.g. \emph{metal nut}) and smaller intra-class distances (e.g. \emph{hazelnut} and \emph{wood}) than those without MGA in the $1^{st}$ row.
In terms of metrics on MVTec AD, our MGA has a 1.5\% NMI gain, 3.2\% ARI gain, and 1.4\% $F_1$ gain shown in Table~\ref{tab:mask_attention}.

We conduct experiments to introduce the binary mask into different tokens in Table~\ref{tab:mask_attention}.
(b) introduces the binary mask on $\texttt{[CLS]}$ token and patch tokens.
(c) introduces it on patch tokens.
Our MGA (d) only performs on $\texttt{[CLS]}$ token, which is directly input into the classifier.
We make patch tokens have a global receptive field to introduce more surrounding information indirectly into \texttt{[CLS]} token.
Compared to other mask strategies, our MGA achieves 5.0\% NMI, 5.9\% ARI, and 2.6\% $F_1$ improvements.

\begin{table}[!t]
    \vspace{-0.5em}
    \centering
    \setlength{\tabcolsep}{2.5mm}
    \resizebox{0.8\linewidth}{!}{
    \begin{tabular}{l|ccc}
        \toprule
        Mask Mechanism & NMI & ARI & $F_1$ \\ 
        \midrule
        (a) w/o MGA & 0.598 & 0.494 & 0.698 \\ 
        (b) all tokens  & 0.507 & 0.382 & 0.600 \\ 
        (c) patch tokens & 0.563 & 0.467 & 0.686 \\ 
        (d) class token (Ours) & \textbf{0.613} & \textbf{0.526} & \textbf{0.712} \\  
        \bottomrule
    \end{tabular}
    }
    \vspace{-0.5em}
    \caption{Compared with different mask attention mechanisms on the MVTec AD dataset, the best-performing result is in bold.}
    \label{tab:mask_attention}
    \vspace{-0.5em}
\end{table}

\begin{table}[!t]
    \centering
    \setlength{\tabcolsep}{2.5mm}
    \resizebox{0.9\linewidth}{!}{
    \begin{tabular}{ccccccc}
        \toprule
        & \multicolumn{3}{c}{MVTec AD}  & \multicolumn{3}{c}{MTD} \\ 
        \cmidrule(l){2-4} \cmidrule(l){5-7} 
         $L_m$ & NMI & ARI & $F_1$ & NMI & ARI & $F_1$ \\ 
        \midrule
        1 & 0.606 & 0.508 & 0.690 & 0.191 & 0.157 & 0.420 \\ 
        3 & 0.608 & 0.511 & 0.694 & 0.209 & 0.173 & 0.436 \\ 
        6 & \textbf{0.613} & 0.519 & \textbf{0.713} & 0.253 & 0.214 & 0.492 \\ 
        9 & \textbf{0.613} & \textbf{0.526} & 0.712 & \textbf{0.268} & \textbf{0.228} & \textbf{0.509}\\
        12 & 0.609 & 0.521 & 0.712 & 0.249 & 0.213 & 0.492 \\
        \bottomrule
    \end{tabular}
    }
    \vspace{-0.5em}
    \caption{Ablation study on the position of mask-guided layers on MVTec AD and MTD datasets. Best results in bold.}
    \label{tab:guide_layer}
    \vspace{-0.5em}
\end{table}

\subsubsection{The influence of the mask-guided layer $L_m$}
In our method, we introduce mask-guided attention in the last $L_m$ layers of ViT.
As reported in Table~\ref{tab:guide_layer}, we conduct the experiments with different $L_m$.
The results demonstrate that using mask-guided attention in the last 9 layers yields better performance on both datasets.
When $L_m$ is small, the $\texttt{[CLS]}$ token cannot adequately focus on the anomaly within the binary mask.
In contrast, when $L_m=12$, the $\texttt{[CLS]}$ token completely lost contextual information around the anomaly,
resulting in a drop in metrics.

\begin{table}[!t]
    \centering
    \setlength{\tabcolsep}{2.5mm}
    \resizebox{1.0\linewidth}{!}{
    \begin{tabular}{lcccccc}
        \toprule
         & \multicolumn{3}{c}{MVTec AD}  & \multicolumn{3}{c}{MTD} \\ 
        \cmidrule(l){2-4} \cmidrule(l){5-7} 
         Merge & NMI & ARI & $F_1$ & NMI & ARI & $F_1$ \\ 
        \midrule
        (\romannumeral1) Avg & 0.610 & 0.521 & 0.709 & 0.257 & 0.223 & 0.501 \\ 
        (\romannumeral2) Score Avg & 0.600 & 0.513 & 0.703 & 0.228 & 0.208 & 0.485 \\ 
        (\romannumeral3) Area Avg & \textbf{0.613} & \textbf{0.526} & \textbf{0.712} & \textbf{0.268} & \textbf{0.228} & \textbf{0.509} \\ 
        \bottomrule
    \end{tabular}
    }
    \vspace{-0.5em}
    \caption{Compared with different merging strategies on the MVTec AD and MTD datasets.}
    \label{tab:merge}
    \vspace{-0.5em}
\end{table}

\subsubsection{Discussion of merging strategies}
In Table~\ref{tab:merge}, we report results using different merging strategies.
(\romannumeral1) averaging the predictions of the sub-images, (\romannumeral2) utilizing the anomaly scores of the sub-images as weights to merge, and (\romannumeral3) using our region merging strategy.
Strategy (\romannumeral3) achieves better metrics than (\romannumeral1) and (\romannumeral2) on both datasets.
In anomaly detection, noise inevitably leads to over-detections, which have small areas but large anomaly scores.
Directly averaging results over all sub-images is inaccurate, especially when many false positive sub-images are involved.
Similarly, using anomaly scores as weights is also inappropriate.
In contrast, our merging method, which assigns weights to multiple sub-image predictions based on area, achieves more robust merged predictions.

\begin{table}[!t]
    \centering
  \setlength{\tabcolsep}{2.5mm}
  \resizebox{0.9\linewidth}{!}{
    \begin{tabular}{c|cc|ccc}
        \toprule
        & Precision & Recall & NMI & ARI & $F_1$ \\ 
        \midrule 
        w/o PLC  & 0.797 & 0.727 & 0.597 & 0.517 & \textbf{0.714} \\ 
        w PLC & \textbf{0.821} & \textbf{0.876} & \textbf{0.613} & \textbf{0.526} & 0.712  \\ 
        \bottomrule
    \end{tabular}
    }
    \vspace{-0.5em}
    \caption{The ablation experiment of pseudo label correction (PLC) on the MVTec AD dataset. We evaluate the precision and recall of the normal class.}
    \label{tab:plc_ablation}
    \vspace{-0.5em}
\end{table}

\subsubsection{Effect of Pseudo Label Correcting}
\label{exp:pseudo_label}
Our Pseudo Label Correcting (PLC) aims to reduce the impact of over-detections from the anomaly detection process.
Although over-detected sub-images are normal, they have large appearance differences, which makes it difficult to correctly group them into the same normal class.
Table~\ref{tab:plc_ablation} presents the precision and recall of the normal class.
After applying PLC, recall improves by 14.9\%,
which means that some misclassified normal sub-images are corrected by our PLC.
Additionally, a 1.6\% improvement in NMI is observed in the multi-class classification results.

\section{Conclusion}
In this paper, we propose a novel multi-class anomaly classification method AnomalyNCD, which is compatible with existing anomaly detection methods.
Firstly, we propose the main element binarization approach to isolate most over-detections and missed detections brought by anomaly detection methods.
Secondly, we crop each anomalous region and leverage mask-guided representation learning to focus the network on the local anomaly and learn discriminative representations.
Finally, our region merging strategy achieves flexible classification at sub-image and image levels during inference.
While the performance of AnomalyNCD is generally influenced by the quality of the  anomaly detection method,
our AnomalyNCD outperforms all the current industrial anomaly clustering methods and NCD methods in the nature scene.

\section{Acknowledgments}
\noindent This work was supported by the National Natural Science Foundation of China under Grant No.62176098. The computation is completed in the HPC Platform of Huazhong University of Science
 and Technology.

{
    \small
    \bibliographystyle{ieeenat_fullname}
    \bibliography{main}
}

\appendix

\maketitlesupplementary

\section*{Overview}
In this appendix, we provide additional descriptions of the following contents:
\begin{itemize}

    \item Training details for novel class discovery (Sec.~\ref{app:ncd_details}).
        
    \item The implementation details about our experiment settings (Sec.~\ref{app:implementation_details}).

    \item The implementation details of our cropping method (Sec.~\ref{app:imple_crop}).

    \item Discussion of the anomaly-centered sub-image cropping (Sec.~\ref{app:crop_discuss}).

    \item Discussion of the vision transformer under a supervised setting (Sec.~\ref{app:sup_vit}).
    
    \item Discussion of the labeled abnormal images (Sec.~\ref{app:label_image}).

    \item Discussion on handling the combined category (Sec.~\ref{sec:combined}).

    \item The computational analysis (Sec.~\ref{app:comput_analy}).

    \item The analysis of model performance using ground truth masks (Sec.~\ref{app:gt_mask}).

    \item Ablation study on the number of threshold index in MEBin (Sec.~\ref{app:sampling}).

    \item The sensitivity analysis of the hyperparameter $\tau$ in MEBin (Sec.~\ref{app:min_stable}).

    \item Sensitivity analysis of threshold 0.5 in PLC (Sec.~\ref{app:plc}).

    \item The binarization results of MEBin (Sec.~\ref{app:bin_map}).

    \item More detailed t-SNE visualizations (Sec.~\ref{app:detailed_tsne}).
    
    \item More detailed quantitative results (Sec.~\ref{app:detailed_mvtec}).

    \item Limitation (Sec.~\ref{app:limit}).

\end{itemize}

\section{Training details for novel class discovery}
\label{app:ncd_details}
We present concrete implementations of the novel class discovery paradigm used in our method.
Following \citeapp{appCVPR2022GCD, appICCV2023SimGCD},
we employ both classification learning and contrastive representation learning to train the network.
For each sub-image $x_{i,k}$, we use a new subscript $j$ to replace $i,k$ for a simple and clear description.
Thus the augmented views of $x_{j}$ are formed as $(\tilde{x}_j,\hat{x}_j)$.
These views are fed into MGViT, denoted as $f(\cdot)$, to extract the $\texttt{[CLS]}$ tokens.

\begin{itemize}
\item For labeled sub-images, the ground truth labels are available.
Therefore, we employ both self-supervised contrastive learning and supervised contrastive learning.
In addition, supervised classification learning is performed by treating the ground truth labels as optimization targets.
\item For unlabeled sub-images, we use self-supervised contrastive representation learning.
The pseudo labels are generated to perform classification learning.
\end{itemize}

\textbf{Contrastive representation learning.}
The $\texttt{[CLS]}$ tokens extracted by MGViT are fed into a three-layer MLP, denoted as $\phi(\cdot)$,
to generate the image features from $\tilde{x}_j$ and $\hat{x}_j$:
$\tilde{\boldsymbol{z}}_j\!=\!\phi(f(\tilde{x}_j))$ and $\hat{\boldsymbol{z}}_j\!=\!\phi(f(\hat{x}_j))$.
For both labeled and unlabeled sub-images, we employ the self-supervised contrastive loss $\mathcal{L}_{\text{rep}}$ as,
\begin{equation}
\mathcal{L}_{\text{rep}}=\frac{1}{|B|} \sum_{j\in B} -\log \frac{\exp (\hat{\boldsymbol{z}}_j^{ \top} \tilde{\boldsymbol{z}}_j / \tau_u)}{\sum_{{n \in \mathcal{N}_j}} \exp (\hat{\boldsymbol{z}}_j^{ \top} \tilde{\boldsymbol{z}}_n / \tau_u)}
\end{equation}
where $B$ denotes a mini-batch.
${\left|B\right|}$ represents the number of image pairs in $B$.
$\mathcal{N}_j$ indexes other image pairs in the batch except $(\hat{x}_j,\tilde{x}_j)$,
and $\tau_u$ is a temperature value.
Similarly, the supervised contrastive loss for labeled data is written as:
\begin{equation}
\!\! \mathcal{L}_{\text {rep}}^\mathbf{l}\!=
\frac{1}{|B^\mathbf{l}|}
\sum_{j \in B^\mathbf{l}} \frac{1}{|\mathcal{P}_j|}
\sum_{p \in \mathcal{P}_j}
-\log \frac{\exp (\hat{\boldsymbol{z}}_j^{\top} \tilde{\boldsymbol{z}}_p / \tau_c)}
{\sum_{n \in \mathcal{N}_j} \exp (\hat{\boldsymbol{z}}_j^{\top} \tilde{\boldsymbol{z}}_n / \tau_c)}
\end{equation}
where $B^\mathbf{l}\subset B$ represents the indexes of labeled images in the batch $B$.
$\mathcal{P}_j$ indexes other images in the batch $B$ which have the same labels as $\hat{x}_j$ and $\tilde{x}_j$.
$\tau_c$ is a temperature value.

\textbf{Classification learning.}
For each image pair $(\hat{x}_j,\tilde{x}_j)$ in a batch,
we feed them into the teacher and student networks respectively,
which have shared MGViT and classification heads with different softmax temperatures.
For labeled sub-images, we convert the ground truth label into the one-hot vector $\hat{y}_j=\tilde{y}_j \in \mathbb{R}^{1 \times (\mathcal{C}^\mathbf{l}+\mathcal{C}^\mathbf{u})}$,
where the entry corresponding to the target class is set to $1$ and all other entries are $0$.
For unlabeled sub-images $\tilde{x}_{j},\hat{x}_{j}$, we generate pseudo labels $\hat{q}_{j},\tilde{q}_{j}$ by the ``teacher'' one, which employs a sharp temperature $\tau_t$ to produce confident predictions.
Specifically, following DINO \citeapp{appiccv2021dino}, 
we compute the logits for two views $(\hat{x}_j,\tilde{x}_j)$ as $\hat{l}_{j}\!=\!\mathcal{H}(f(\hat{x}_{j}))$
and $\tilde{l}_{j}\!=\!\mathcal{H}(f(\tilde{x}_{j}))$,
where $\hat{l}_{j}, \tilde{l}_{j} \in \mathbb{R}^{1 \times (\mathcal{C}^\mathbf{l}+\mathcal{C}^\mathbf{u})}$,
$f(\cdot)$ indicates MGViT that outputs the $\texttt{[CLS]}$ token,
and $\mathcal{H}(\cdot)$ is the linear classifier.
Then,
a softmax with temperature $\tau_t$ converts these logits into pseudo labels $\hat{q}_{j},\tilde{q}_{j} \in \mathbb{R}^{1 \times (\mathcal{C}^\mathbf{l}+\mathcal{C}^\mathbf{u})}$ as follows,
\begin{equation}
\hat{q}_{j}^{(k)} = \frac{\exp(\hat{l}_{j}^{(k)}/{\tau_t})}{\sum_{k=1}^{\mathcal{C}^\mathbf{l}+\mathcal{C}^\mathbf{u}}\exp(\hat{l}_{j}^{(k)}/{\tau_t})}
\label{equ:pseudo_q}
\end{equation}
where $\hat{q}_{j}^{(k)}$ and $\hat{l}_{j}^{(k)}$ denote the pseudo-label value and logit value respectively for the $k$-th class in $ \mathcal{C}^\mathbf{l}+\mathcal{C}^\mathbf{u}$.
Note that, we set the logits $\tilde{l}_{j},\hat{l}_{j}$ for $\mathcal{C}^\mathbf{l}$ known classes to $-\infty$ to isolate known classes when calculating pseudo labels.
Then the pseudo-label values for the known classes calculated by Eq. \eqref{equ:pseudo_q} are $0$.
The ``student'' one, using a smooth temperature $\tau_s$, generates the probability predictions $\hat{p}_{j} \in \mathbb{R}^{1 \times (\mathcal{C}^\mathbf{l}+\mathcal{C}^\mathbf{u})}$ for sub-image $\hat{x}_{j}$ as,
\begin{equation}
\hat{p}_{j}^{(k)} = \frac{\exp(\hat{l}_{j}^{(k)}/{\tau_s})}{\sum_{k=1}^{\mathcal{C}^\mathbf{l}+\mathcal{C}^\mathbf{u}}\exp(\hat{l}_{j}^{(k)}/{\tau_s})}
\end{equation}
where $\hat{p}_{j}^{(k)}$ denotes the prediction probability for the $k$-th class in $\mathcal{C}^\mathbf{l}+\mathcal{C}^\mathbf{u}$.
The student network is trained to align $\hat{p}_{j}$, $\tilde{p}_{j}$ with either the ground truth labels (for labeled data) or the pseudo labels (for unlabeled data).
With the representation above,
we leverage the standard cross-entropy loss $\mathcal{L}_{\text{CE}}(p,q)\!=\!-\sum_{c=0}^{\mathcal{C}^\mathbf{l}+\mathcal{C}^\mathbf{u}-1}p^{c}\text{log}q^{c}$ to optimize the classification, which is represented as:
\begin{align}
\mathcal{L}_{\text{cls}}^\mathbf{l} &= \frac{1}{\left|B^\mathbf{l}\right|} \sum\nolimits_{j \in B^\mathbf{l}} (\mathcal{L}_{\text{CE}}(\hat{y}_j, \tilde{p}_j)+\mathcal{L}_{\text{CE}}(\tilde{y}_j, \hat{p}_j)) \\
\mathcal{L}_{\text{cls}}^\mathbf{u} &= \frac{1}{\left|B^\mathbf{u}\right|} \sum\nolimits_{j \in B^\mathbf{u}} (\mathcal{L}_{\text{CE}}(\hat{q}_j, \tilde{p}_j)+\mathcal{L}_{\text{CE}}(\tilde{q}_j, \hat{p}_j))
\end{align}
where $B^\mathbf{u}$ indexes the unlabeled images of the batch $B$.

Finally, we adopt a mean-entropy maximization regularizer \citeapp{appNIPS2020entropy1}
$\mathcal{L}_{\text{reg}}^\mathbf{u} = \mathcal{L}_{\text{CE}}(\bar{p}_j,\bar{p}_j)$ for unlabeled sub-images,
where $\bar{p}_j=\frac{1}{2\left|B^\mathbf{u}\right|} \sum_{j \in B^\mathbf{u}}(\hat{p}_j+\tilde{p}_j)$ denotes the mean prediction of unlabeled images.
The overall loss of training objective is written as:
\begin{equation}
\mathcal{L} = \lambda(\mathcal{L}_{\text {rep}}^\mathbf{l}+\mathcal{L}_{\text{cls}}^\mathbf{l}) +(1-\lambda)(\mathcal{L}_{\text {rep}} + \mathcal{L}_{\text{cls}}^\mathbf{u}+\mu \mathcal{L}_{\text{reg}}^\mathbf{u})
\end{equation}
where $\lambda$ is a hyperparameter that balances the supervised loss and the self-supervised loss.
$\mu$ is the coefficient of the regularization.

\section{Additional implementation details}
\label{app:implementation_details}
We use ViT-B/8 \citeapp{appICLR2021ViT} pre-trained with DINO \citeapp{appiccv2021dino} as our feature extractor.
The self-attentions in the last 9 layers are replaced with our mask-guided attention. 
During training, all the layers of ViT are fixed except the last layer.
As with UNO\citeapp{appICCV2021UNO} and BYOP\citeapp{appCVPR2023BYOP}, we use the multi-head strategy in the classifier $\mathcal{H}$,
and use the head with the smallest loss for inference.
The input images are scaled to a resolution of $224 \times 224$.
We set the batch size to 32 and epochs to 50.
The Stochastic Gradient Descent (SGD) optimizer \citeapp{appSGD} is employed with a learning rate of 0.003.
To generate two augmented views for contrastive learning,
we follow the data augmentation strategies in BYOL \citeapp{appNIPS2020BYOL} (random crop, flip, color jittering, and Gaussian blur) and RandAug \citeapp{appCVPRW2020RandAug} (rotation, posterize, and sharpness).
Aligning with \citeapp{appICCV2023SimGCD},
the temperature values are set as follows: $\tau_u$ is 0.07 and $\tau_c$ is 1.0, $\tau_s$ is set to 0.1 and $\tau_t$ is initialized to 0.07.
For the first 40 epochs, $\tau_t$ reduces every 4 epochs, linearly down to 0.04, then stays the same.
For our Main Element Binarization approach, we set the $\mathcal{T}$ to 64 and $\tau$ to 4.
In the overall loss, the parameters $\lambda$ and $\mu$ are set to 0.3 and 4 respectively.
In the Region Merging strategy, the temperature $\tau_\alpha$ is set to 100 on the MVTec AD dataset and 50 on the MTD dataset.

For anomaly detection methods preceding AnomalyNCD, we choose zero-shot method MuSc \citeapp{appICLR2024MuSc}, one-class methods PatchCore \citeapp{appCVPR2022patchcore}, EfficientAD \citeapp{appwacv2024efficientad}, RD++ \citeapp{appcvpr2023rd}, PNI \citeapp{appiccv2023pni} and CPR \citeapp{appTIP2024CPR}.
Note that CPR does not provide the training code, we directly use the official checkpoints on the MVTec AD dataset.
EfficientAD does not release official implementation, we use the unofficial version.
For RD++ and PNI, we conduct experiments on the MTD dataset according to the configuration of MVTec AD.

\begin{table*}[!t]
    \centering
    \setlength{\tabcolsep}{2.2mm}
    \resizebox{1.0\linewidth}{!}{
    \begin{tabular}{llcccccccccccc}
        \toprule
        & & \multicolumn{3}{c}{MuSc\citeapp{appICLR2024MuSc}+AnomalyNCD} & \multicolumn{3}{c}{PatchCore\citeapp{appCVPR2022patchcore}+AnomalyNCD} & \multicolumn{3}{c}{EfficientAD\citeapp{appwacv2024efficientad}+AnomalyNCD} & \multicolumn{3}{c}{PNI\citeapp{appiccv2023pni}+AnomalyNCD} \\ 
        \cmidrule(l){3-5} \cmidrule(l){6-8} \cmidrule(l){9-11} \cmidrule(l){12-14} 
        Dataset & Setting & NMI & ARI & $F_1$ & NMI & ARI & $F_1$ & NMI & ARI & $F_1$ & NMI & ARI & $F_1$ \\ 
        \midrule
        \multirow{2}{*}{MVTec AD \citeapp{appCVPR2019mvtec}} & w/o crop & \textbf{0.622} & \textbf{0.549} & \textbf{0.740} & 0.650 & 0.584 & 0.755 & 0.501 & 0.396 & 0.637 & \textbf{0.676} & \textbf{0.616} & \textbf{0.777} \\ 
        & w crop & 0.613 & 0.526 & 0.712 & \textbf{0.670} & \textbf{0.601} & \textbf{0.769} & \textbf{0.516} & \textbf{0.394} & \textbf{0.641} & 0.675 & 0.609 & 0.769 \\ 
        \midrule
        \multirow{2}{*}{MTD \citeapp{appVC2020MTD}} & w/o crop & 0.086 & 0.076 & 0.374 & 0.142 & 0.166 & 0.432 & 0.105 & 0.066 & 0.379 & 0.107 & 0.134 & 0.407 \\ 
        & w crop & \textbf{0.268} & \textbf{0.228} & \textbf{0.509} & \textbf{0.380} & \textbf{0.390} & \textbf{0.617} & \textbf{0.220} & \textbf{0.188} & \textbf{0.467} & \textbf{0.181} & \textbf{0.219} & \textbf{0.465} \\
        \bottomrule
    \end{tabular}
    }
    \caption{Ablation of Anomaly-Centered Sub-Image Cropping on the MVTec AD and MTD datasets.}
    \label{tab:crop}
\end{table*}

\section{Implementation details of cropping}
\label{app:imple_crop}
For each region in $\mathbf{M}_i^\mathbf{u}$,
we set a square box that completely encloses this region but has minimal area.
Then we extend each square box by adding a 10\% padding.
The padding area includes the background of anomalies and thus enables our network to extract features indicating the anomaly's position on products, which is critical for classification.
Notice that we keep the minimum crop size as 1\% of the image size.
Finally, along each square box of $\mathbf{M}_i^\mathbf{u}$,
we crop a sub-image from $I_i^\mathbf{u}$ and its mask from $\mathbf{M}_i^\mathbf{u}$.

\begin{figure}[!t]
\centering
\includegraphics[width=0.48\textwidth]{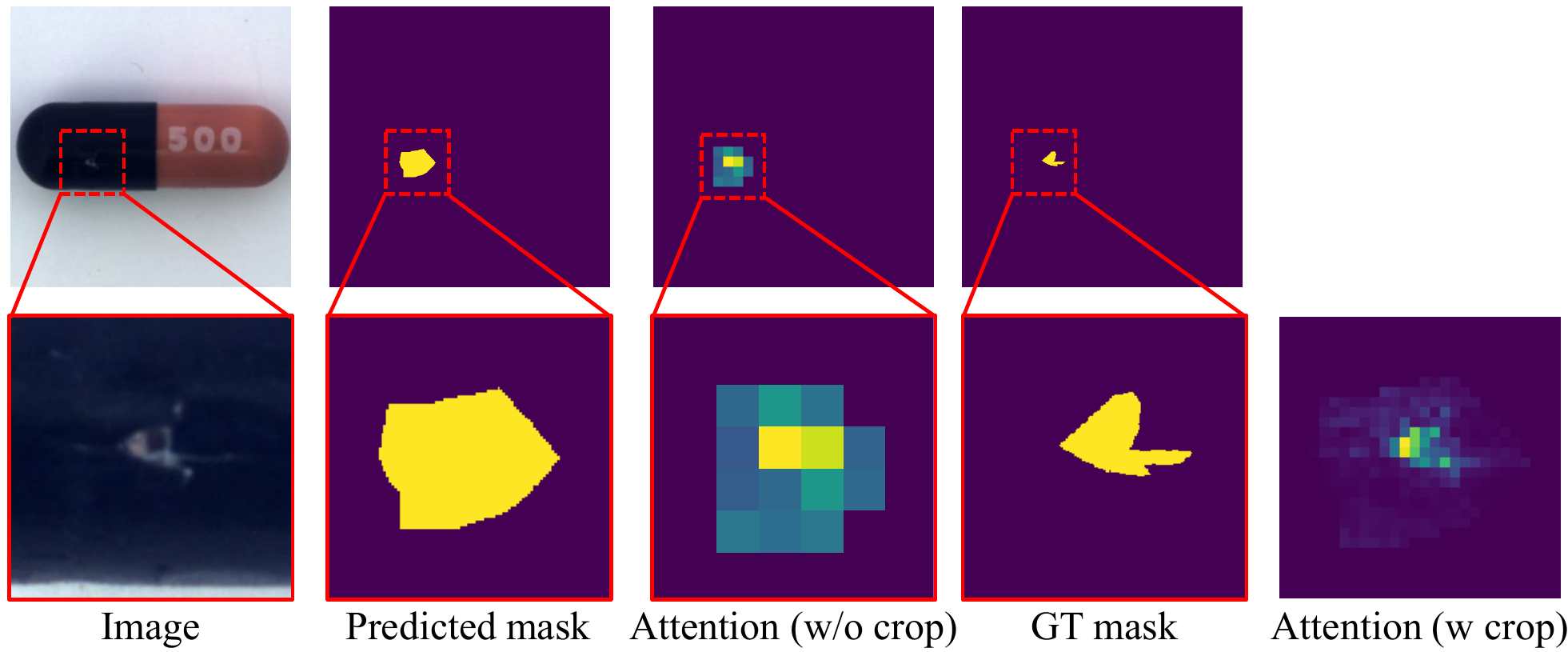}
\caption{\textbf{Visualization of $\texttt{[CLS]}$ tokens's attention} for entire image and sub-image.}
\label{no_crop_attn}
\end{figure}

\section{Discussion of the anomaly-centered sub-image cropping}
\label{app:crop_discuss}

The cropping operation plays a crucial role in industrial anomaly clustering for two primary reasons.
First, many anomalies in industrial products are local and subtle.
The cropping operation makes the anomalous region occupy most of the image, which helps the network learn the anomaly easily.
Second, the cropping operation can handle images with \textit{combined} type anomalies, where there are different types of anomalies in an image. 
More details about \textit{combined} type anomalies are given in Sec.~\ref{sec:combined}.

As shown in Fig.~\ref{no_crop_attn}, for the original image,
the attention of $\texttt{[CLS]}$ token has a coarse-grained response in the anomalous region.
When the anomaly is subtle, the discriminative features are hard to learn.
For the sub-image, the attention has a fine-grained response and learns the more discriminative pixels of subtle anomaly.
We report multi-class classification on the MVTec AD and MTD datasets in Table~\ref{tab:crop}.
Since anomalies on the MTD dataset are subtle and finer, the crop operation is required for fine-grained feature learning,
which can bring up to 23.8\% NMI improvement for PatchCore+AnomalyNCD and 11.5\% NMI gains for EfficientAD+AnomalyNCD.
On the MVTec AD dataset, the crop operation results in a 2.0\% NMI increase for PatchCore+AnomalyNCD and 1.5\% NMI gain for EfficientAD+AnomalyNCD.
When using MuSc or PNI as the anomaly detection method, it only decreases NMI by 0.9\% at most.

\begin{figure}[!t]
    \centering
    \begin{minipage}{0.49\linewidth}
        \centering
        \includegraphics[width=\textwidth]{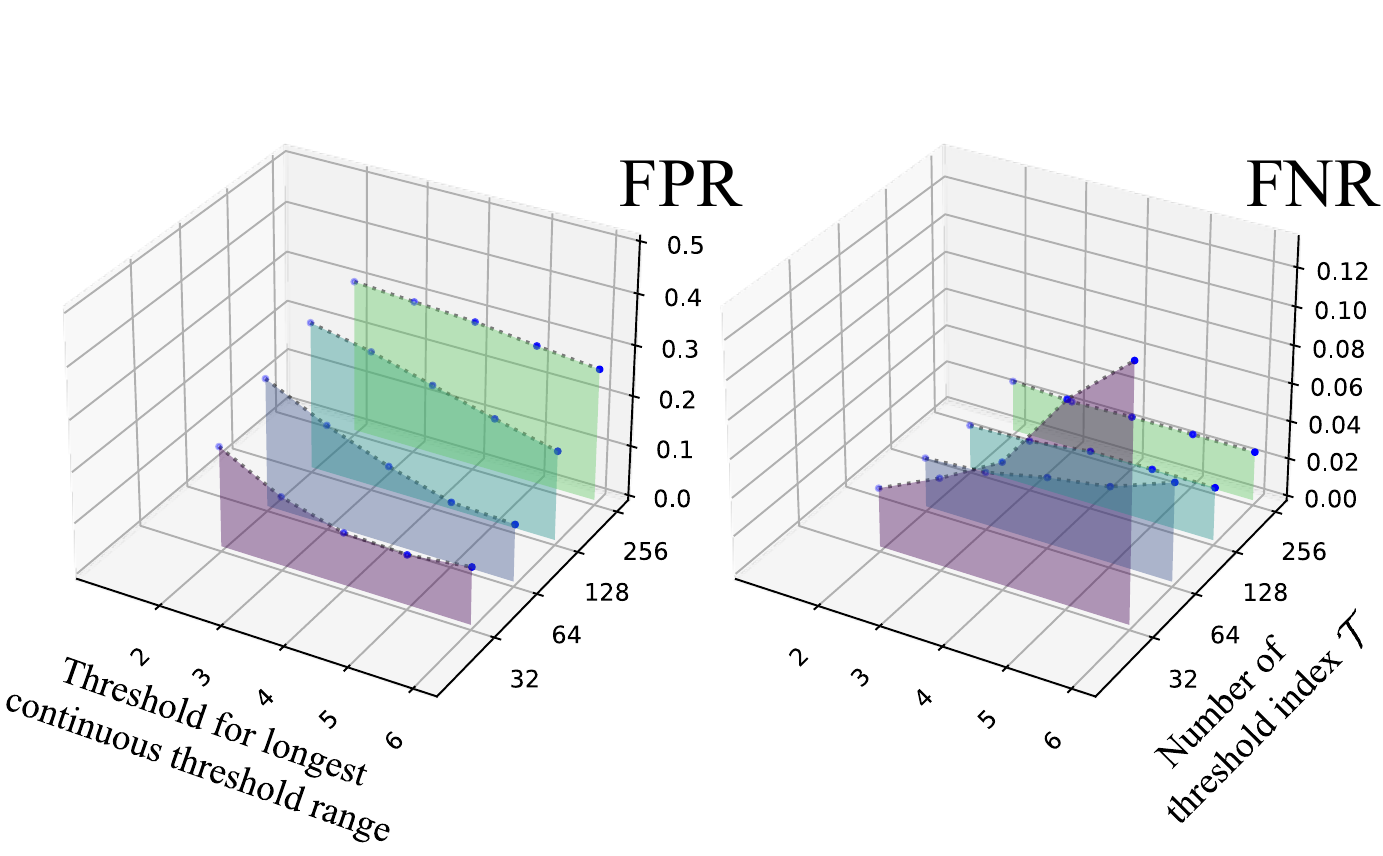}
        \caption{Ablation study on $\mathcal{T}$ in MEBin.}
        \label{fig:mebin}
    \end{minipage}
    \hfill
    \begin{minipage}{0.49\linewidth}
        \centering
        \includegraphics[width=\textwidth]{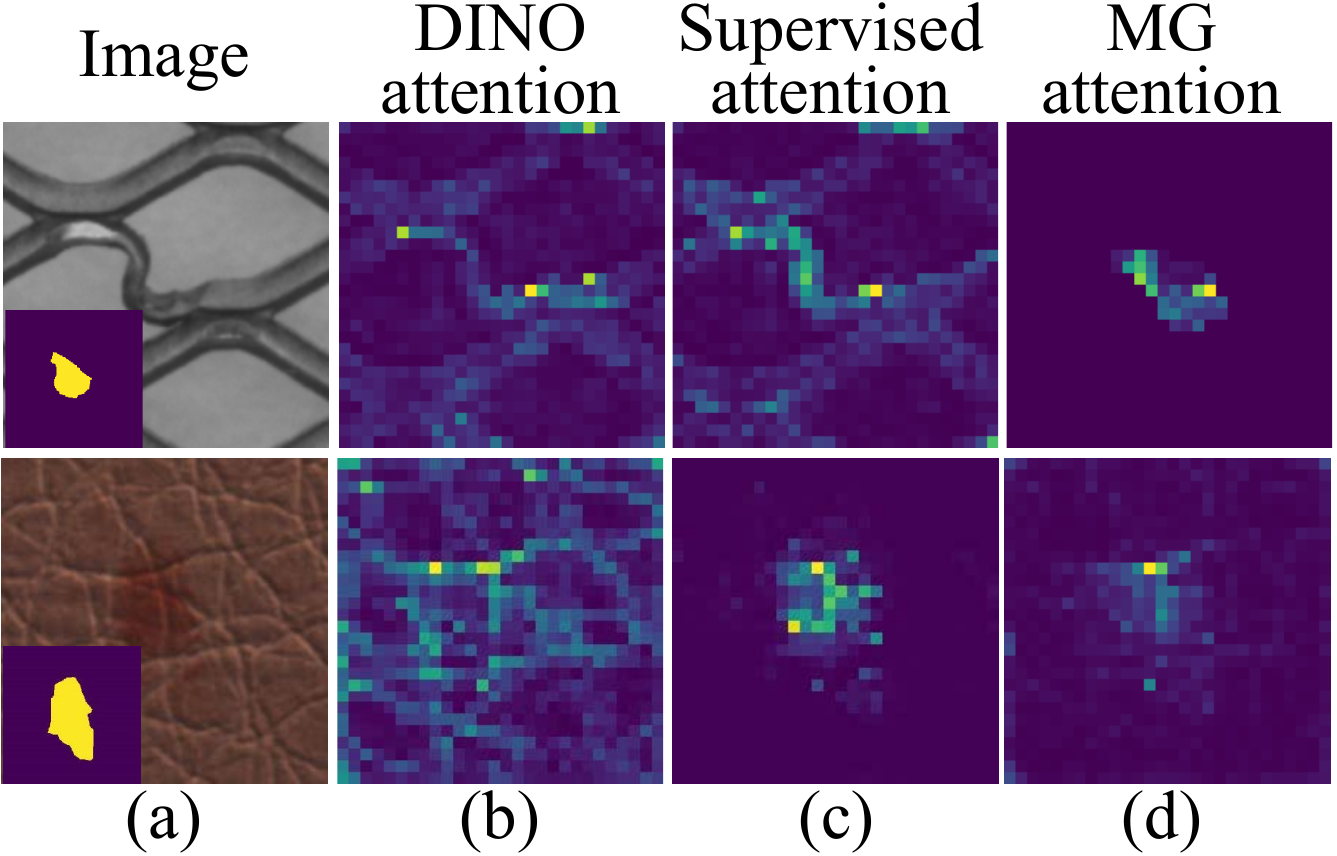}
        \caption{Visualization of self-attention of the \texttt{[CLS]} token.}
        \label{fig:sup_attn}
    \end{minipage}
\vspace{-0.1cm}
\end{figure}

\section{Discussion of ViT under a supervised setting}
\label{app:sup_vit}
To study the ability of the vision transformer (ViT) to focus on local anomalies under a supervised setting, we leave one out cross validation.
As shown in the attention maps of Fig. \ref{fig:sup_attn},
the supervised training could focus on the anomaly regions, while it needs manually collecting and labeling the abnormal samples, which is time-consuming.
As shown in (c) and (d), our AnomalyNCD leverages self-supervision to reduce these manual operations while also focuses attention on the anomalies,
which is more suitable for industrial scenarios.

\section{Discussion of the labeled abnormal images}
\label{app:label_image}

\begin{figure}[t]
\vspace{-0.5em}
\begin{center}
\includegraphics[width=0.49\textwidth]{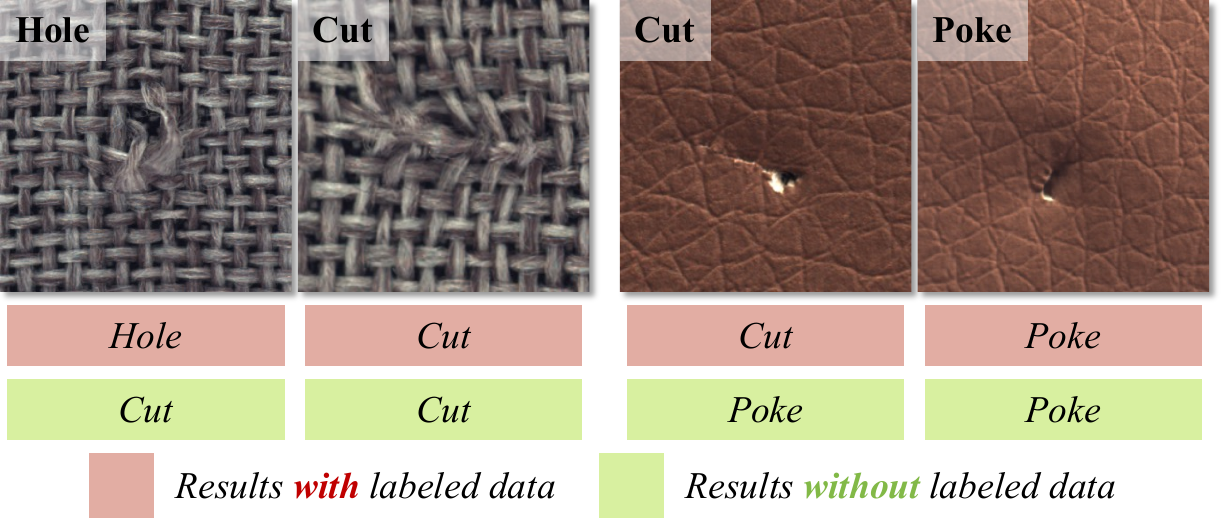}
\caption{\textbf{Impact of using or not using labeled data}.
We show two product cases, and each presents two similar anomalies.}
\vspace{-8pt}
\label{fig:label_effect}
\end{center}
\vspace{-0.5em}
\end{figure}

As shown in Fig.~\ref{fig:label_effect},
labeled data is significant for our mask-guided representation learning (MGRL).
If only using unlabeled data,
MGRL is prone to confuse anomalies with highly similar appearances.
For example, a ``Hole'' on a carpet is treated as ``Cut'', and ``Cut'' on leather is misclassified as a ``Poke''.
Actually, even humans tend to confuse them.
With the supervision of labeled data,
confusion does not occur.
The reason is that the labeled data helps to build the feature space of prior classes,
so that even the subtle differences from novel classes can be recognized.

In the labeled abnormal images $\mathcal{D}^\mathbf{l}$, there is much prior knowledge of industrial anomalies classification, such as anomalies with similar size, color, and location belonging to the same class.
Using them to train the network together can transfer the knowledge from $\mathcal{D}^\mathbf{l}$ to the network and separate confusing anomalies from each other in $\mathcal{D}^\mathbf{u}$.
We report the quantitative results in Table~\ref{tab:base_class}, using $\mathcal{D}^\mathbf{l}$ brings 3.0\% NMI improvements on MVTec AD and 3.9 \% NMI improvements on MTD.
In addition, we also use 15 categories in the MVTec AD dataset as the labeled abnormal images in turn.
Except for the \emph{toothbrush} category, which has only one anomaly class.
The quantitative results are reported in Tables~\ref{tab:mvtec_base_nmi} to \ref{tab:mvtec_base_f1}.
MuSc is used as the anomaly detection method by default.

\begin{table}[!t]
    \centering
  \setlength{\tabcolsep}{2.5mm}
  \resizebox{1.0\linewidth}{!}{
    \begin{tabular}{ccccccc}
        \toprule
        & \multicolumn{3}{c}{MVTec AD}  & \multicolumn{3}{c}{MTD} \\ 
        \cmidrule(l){2-4} \cmidrule(l){5-7} 
        & NMI & ARI & $F_1$ & NMI & ARI & $F_1$ \\ 
        \midrule 
        w/o $\mathcal{D}^\mathbf{l}$  & 0.583 & 0.506 & 0.689 & 0.227 & 0.202 & 0.485 \\ 
        w $\mathcal{D}^\mathbf{l}$ & \textbf{0.613} & \textbf{0.526} & \textbf{0.712} & \textbf{0.268} & \textbf{0.228} & \textbf{0.509} \\ 
        \bottomrule
    \end{tabular}
    }
    \caption{The ablation experiment of using labeled abnormal images $\mathcal{D}^\mathbf{l}$ on the MVTec AD and MTD datasets.}
  \label{tab:base_class}
\end{table}

\begin{table*}[h]
    \centering
  \setlength{\tabcolsep}{1mm}
  \resizebox{1.0\linewidth}{!}{
    \begin{tabular}{cccccccccccccccc|c}
        \toprule
        train~\textbackslash~test& bottle & cable & capsule & carpet & grid & hazelnut & leather & metal\_nut & pill & screw & tile & toothbrush & transistor & wood & zipper & mean\\ 
        \midrule 
        bottle & - & 0.561 & 0.483 & 0.840 & 0.593 & 0.727 & 0.623 & 0.744 & 0.434 & 0.339 & 0.867 & 0.269 & 0.535 & 0.731 & 0.442 & 0.585 \\ 
        cable & 0.676 & - & 0.522 & 0.862 & 0.699 & 0.695 & 0.784 & 0.799 & 0.396 & 0.435 & 0.858 & 0.341 & 0.511 & 0.640 & 0.545 & 0.626 \\ 
        capsule & 0.614 & 0.604 & - & 0.714 & 0.652 & 0.715 & 0.724 & 0.674 & 0.412 & 0.433 & 0.719 & 0.242 & 0.498 & 0.702 & 0.586 & 0.603 \\ 
        carpet & 0.564 & 0.571 & 0.503 & - & 0.624 & 0.714 & 0.818 & 0.745 & 0.416 & 0.418 & 0.885 & 0.218 & 0.508 & 0.681 & 0.541 & 0.586 \\ 
        grid & 0.552 & 0.520 & 0.460 & 0.790 & - & 0.747 & 0.714 & 0.707 & 0.441 & 0.361 & 0.794 & 0.218 & 0.517 & 0.680 & 0.555 & 0.575 \\ 
        hazelnut & 0.535 & 0.616 & 0.451 & 0.731 & 0.660 & - & 0.640 & 0.799 & 0.401 & 0.351 & 0.824 & 0.299 & 0.513 & 0.712 & 0.493 & 0.573 \\ 
        leather & 0.639 & 0.603 & 0.443 & 0.755 & 0.597 & 0.740 & - & 0.658 & 0.439 & 0.419 & 0.885 & 0.368 & 0.507 & 0.663 & 0.570 & 0.592 \\ 
        metal\_nut & 0.598 & 0.556 & 0.575 & 0.863 & 0.723 & 0.684 & 0.838 & - & 0.463 & 0.355 & 0.811 & 0.242 & 0.521 & 0.617 & 0.559 & 0.600 \\ 
        pill & 0.603 & 0.594 & 0.430 & 0.850 & 0.641 & 0.753 & 0.763 & 0.728 & - & 0.398 & 0.885 & 0.368 & 0.543 & 0.678 & 0.518 & 0.625 \\ 
        screw & 0.569 & 0.559 & 0.486 & 0.746 & 0.501 & 0.688 & 0.815 & 0.787 & 0.433 & - & 0.895 & 0.368 & 0.493 & 0.685 & 0.542 & 0.612 \\ 
        tile & 0.514 & 0.587 & 0.506 & 0.762 & 0.501 & 0.750 & 0.790 & 0.725 & 0.425 & 0.414 & - & 0.242 & 0.483 & 0.693 & 0.548 & 0.565 \\ 
        transistor & 0.499 & 0.607 & 0.420 & 0.645 & 0.569 & 0.699 & 0.651 & 0.768 & 0.446 & 0.387 & 0.738 & 0.299 & - & 0.445 & 0.553 & 0.552 \\ 
        wood & 0.586 & 0.633 & 0.472 & 0.764 & 0.655 & 0.708 & 0.700 & 0.789 & 0.402 & 0.354 & 0.767 & 0.218 & 0.482 & - & 0.533 & 0.576 \\ 
        zipper & 0.627 & 0.555 & 0.395 & 0.795 & 0.616 & 0.723 & 0.765 & 0.755 & 0.432 & 0.412 & 0.803 & 0.398 & 0.543 & 0.658 & - & 0.606 \\ 
        \midrule
        mean & 0.583 & 0.582 & 0.473 & 0.778 & 0.618 & 0.719 & 0.740 & 0.744 & 0.426 & 0.390 & 0.825 & 0.292 & 0.512 & 0.660 & 0.537 & 0.591 \\
        \bottomrule
    \end{tabular}
    }
    \caption{The detailed NMI metric evaluated on different labeled abnormal images $\mathcal{D}^\mathbf{l}$ on the MVTec AD dataset.}
  \label{tab:mvtec_base_nmi}
\end{table*}

\begin{table*}[h]
    \centering
  \setlength{\tabcolsep}{1mm}
  \resizebox{1.0\linewidth}{!}{
    \begin{tabular}{cccccccccccccccc|c}
        \toprule
        train~\textbackslash~test& bottle & cable & capsule & carpet & grid & hazelnut & leather & metal\_nut & pill & screw & tile & toothbrush & transistor & wood & zipper & mean\\ 
        \midrule 
        bottle & - & 0.608 & 0.402 & 0.778 & 0.457 & 0.704 & 0.569 & 0.623 & 0.325 & 0.242 & 0.816 & 0.259 & 0.442 & 0.660 & 0.322 & 0.515 \\ 
        cable & 0.643 & - & 0.384 & 0.783 & 0.625 & 0.712 & 0.689 & 0.659 & 0.267 & 0.297 & 0.835 & 0.210 & 0.474 & 0.478 & 0.404 & 0.533 \\ 
        capsule & 0.554 & 0.628 & - & 0.600 & 0.585 & 0.705 & 0.648 & 0.544 & 0.302 & 0.320 & 0.575 & 0.211 & 0.411 & 0.633 & 0.445 & 0.517 \\ 
        carpet & 0.495 & 0.602 & 0.352 & - & 0.546 & 0.715 & 0.789 & 0.606 & 0.289 & 0.315 & 0.830 & 0.167 & 0.460 & 0.599 & 0.409 & 0.512 \\ 
        grid & 0.532 & 0.580 & 0.313 & 0.696 & - & 0.744 & 0.672 & 0.563 & 0.318 & 0.257 & 0.744 & 0.167 & 0.471 & 0.555 & 0.430 & 0.503 \\ 
        hazelnut & 0.474 & 0.643 & 0.361 & 0.679 & 0.610 & - & 0.581 & 0.659 & 0.262 & 0.257 & 0.781 & 0.312 & 0.430 & 0.617 & 0.378 & 0.503 \\ 
        leather & 0.586 & 0.643 & 0.295 & 0.688 & 0.533 & 0.717 & - & 0.515 & 0.298 & 0.310 & 0.830 & 0.259 & 0.407 & 0.569 & 0.416 & 0.505 \\ 
        metal\_nut & 0.562 & 0.482 & 0.452 & 0.789 & 0.664 & 0.702 & 0.765 & - & 0.313 & 0.301 & 0.773 & 0.211 & 0.428 & 0.486 & 0.422 & 0.525 \\ 
        pill & 0.562 & 0.501 & 0.303 & 0.791 & 0.549 & 0.763 & 0.671 & 0.573 & - & 0.273 & 0.830 & 0.259 & 0.464 & 0.608 & 0.395 & 0.539 \\ 
        screw & 0.519 & 0.618 & 0.334 & 0.707 & 0.394 & 0.707 & 0.743 & 0.675 & 0.286 & - & 0.849 & 0.259 & 0.416 & 0.635 & 0.437 & 0.541 \\ 
        tile & 0.478 & 0.635 & 0.367 & 0.697 & 0.420 & 0.745 & 0.679 & 0.570 & 0.278 & 0.322 & - & 0.211 & 0.380 & 0.570 & 0.429 & 0.477 \\ 
        transistor & 0.467 & 0.544 & 0.313 & 0.551 & 0.496 & 0.700 & 0.588 & 0.640 & 0.296 & 0.301 & 0.670 & 0.312 & - & 0.347 & 0.401 & 0.473 \\ 
        wood & 0.539 & 0.639 & 0.370 & 0.684 & 0.599 & 0.719 & 0.620 & 0.656 & 0.285 & 0.232 & 0.732 & 0.167 & 0.416 & - & 0.395 & 0.504 \\ 
        zipper & 0.571 & 0.476 & 0.268 & 0.769 & 0.566 & 0.740 & 0.726 & 0.629 & 0.274 & 0.286 & 0.747 & 0.313 & 0.455 & 0.585 & - & 0.529 \\ 
        \midrule 
        mean & 0.537 & 0.585 & 0.347 & 0.709 & 0.542 & 0.721 & 0.672 & 0.609 & 0.292 & 0.286 & 0.770 & 0.237 & 0.435 & 0.565 & 0.406 & 0.513 \\
        \bottomrule
    \end{tabular}
    }
    \caption{The detailed ARI metric evaluated on different labeled abnormal images $\mathcal{D}^\mathbf{l}$ on the MVTec AD dataset.}
  \label{tab:mvtec_base_ari}
\end{table*}

\begin{table*}[b]
    \centering
  \setlength{\tabcolsep}{1mm}
  \resizebox{1.0\linewidth}{!}{
    \begin{tabular}{cccccccccccccccc|c}
        \toprule
        train~\textbackslash~test& bottle & cable & capsule & carpet & grid & hazelnut & leather & metal\_nut & pill & screw & tile & toothbrush & transistor & wood & zipper & mean\\ 
        \midrule 
        bottle & - & 0.655 & 0.636 & 0.821 & 0.641 & 0.809 & 0.661 & 0.678 & 0.500 & 0.450 & 0.923 & 0.762 & 0.640 & 0.853 & 0.489 & 0.680 \\ 
        cable & 0.843 & - & 0.636 & 0.812 & 0.795 & 0.827 & 0.766 & 0.687 & 0.460 & 0.513 & 0.932 & 0.738 & 0.620 & 0.662 & 0.600 & 0.706 \\ 
        capsule & 0.795 & 0.676 & - & 0.709 & 0.718 & 0.782 & 0.694 & 0.617 & 0.480 & 0.513 & 0.740 & 0.738 & 0.630 & 0.838 & 0.681 & 0.702 \\ 
        carpet & 0.747 & 0.640 & 0.644 & - & 0.705 & 0.782 & 0.847 & 0.661 & 0.453 & 0.506 & 0.932 & 0.714 & 0.630 & 0.824 & 0.607 & 0.692 \\ 
        grid & 0.795 & 0.640 & 0.614 & 0.778 & - & 0.818 & 0.774 & 0.617 & 0.507 & 0.456 & 0.855 & 0.714 & 0.660 & 0.794 & 0.652 & 0.691 \\ 
        hazelnut & 0.699 & 0.705 & 0.636 & 0.778 & 0.782 & - & 0.694 & 0.687 & 0.453 & 0.475 & 0.889 & 0.786 & 0.620 & 0.829 & 0.533 & 0.683 \\ 
        leather & 0.807 & 0.676 & 0.553 & 0.795 & 0.679 & 0.791 & - & 0.617 & 0.487 & 0.525 & 0.932 & 0.762 & 0.620 & 0.809 & 0.644 & 0.693 \\ 
        metal\_nut & 0.807 & 0.619 & 0.652 & 0.803 & 0.833 & 0.800 & 0.798 & - & 0.533 & 0.513 & 0.880 & 0.738 & 0.630 & 0.647 & 0.622 & 0.705 \\ 
        pill & 0.807 & 0.626 & 0.576 & 0.838 & 0.718 & 0.864 & 0.742 & 0.643 & - & 0.488 & 0.932 & 0.762 & 0.620 & 0.838 & 0.600 & 0.718 \\ 
        screw & 0.783 & 0.647 & 0.652 & 0.795 & 0.577 & 0.773 & 0.774 & 0.704 & 0.507 & - & 0.940 & 0.762 & 0.620 & 0.838 & 0.652 & 0.716 \\ 
        tile & 0.759 & 0.691 & 0.667 & 0.735 & 0.564 & 0.818 & 0.750 & 0.617 & 0.447 & 0.506 & - & 0.738 & 0.600 & 0.794 & 0.645 & 0.665 \\ 
        transistor & 0.675 & 0.647 & 0.591 & 0.658 & 0.667 & 0.800 & 0.702 & 0.678 & 0.513 & 0.506 & 0.786 & 0.786 & - & 0.618 & 0.637 & 0.662 \\ 
        wood & 0.795 & 0.683 & 0.629 & 0.761 & 0.769 & 0.800 & 0.702 & 0.687 & 0.493 & 0.444 & 0.855 & 0.714 & 0.630 & - & 0.607 & 0.684 \\ 
        zipper & 0.807 & 0.633 & 0.545 & 0.846 & 0.705 & 0.836 & 0.823 & 0.687 & 0.467 & 0.494 & 0.880 & 0.786 & 0.630 & 0.824 & - & 0.712 \\ 
        \midrule 
        mean & 0.778 & 0.657 & 0.618 & 0.779 & 0.704 & 0.808 & 0.748 & 0.660 & 0.485 & 0.491 & 0.883 & 0.750 & 0.627 & 0.782 & 0.613 & 0.694 \\
        \bottomrule
    \end{tabular}
    }
    \caption{The detailed $F_1$ metric evaluated on different labeled abnormal images $\mathcal{D}^\mathbf{l}$ on the MVTec AD dataset.}
  \label{tab:mvtec_base_f1}
\end{table*}

\begin{figure*}[t]
\centering
\includegraphics[width=1.0\textwidth]{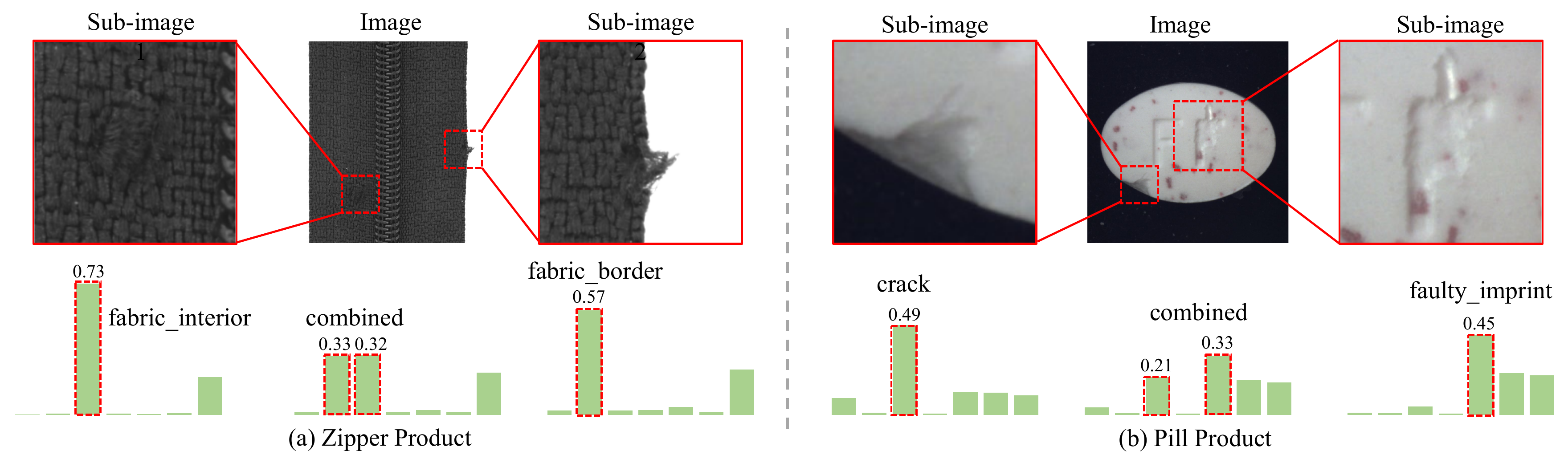}
\caption{
\textbf{Multi-class classification results of two combined-type anomaly images.}
Taking ``zipper'' and ``pill'' as examples,
we show the predicted probabilities for each sub-image individually and for the entire image.
}
\label{abl_combined}
\end{figure*}

\section{Discussion on handling the combined category}
\label{sec:combined}
In main experiments, we follow Anomaly Clustering \citeapp{appWACV2023AC} to remove the combined class for a fair comparison.
However, the combined class is quite common in the industrial scene, such as the \emph{cable}, \emph{pill}, \emph{wood}, and \emph{zipper} products on the MVTec AD dataset.
The image with the combined class contains multiple different types of anomalies,
yet current clustering methods can only classify the entire image into one anomaly type.
For our method, we crop the image into many sub-images, and the classifier assigns each sub-image a label.
In this way, all the anomalous regions in the combined image can be found and classified individually.
As shown in Fig.~\ref{abl_combined}, there are two types of anomalies in the zipper and pill products respectively.
Each anomalous region is cropped and correctly classified.
By merging the prediction of the sub-images to the entire image, we find that classifying the image as these two types of anomalies has a higher probability.

\begin{table}[!t]
    \centering
  \setlength{\tabcolsep}{2.5mm}
  \resizebox{0.75\linewidth}{!}{
    \begin{tabular}{c|cccc}
        \toprule
        & (\uppercase\expandafter{\romannumeral2}) & (\uppercase\expandafter{\romannumeral3}) & (\uppercase\expandafter{\romannumeral4}) & Total\\ 
        \midrule 
        Time (ms)  & 129.8 & 22.8 & 0.5 & 153.1\\ 
        \bottomrule
    \end{tabular}
    }
    \vspace{-0.5em}
    \caption{Inference time for each module on a per-image basis.}
    \label{tab:inference_time}
    \vspace{-0.5em}
\end{table}

\section{Computational analysis}
\label{app:comput_analy}
We measure the inference speed of AnomalyNCD on an NVIDIA RTX 3090 GPU.
The entire inference process consists of four steps: 
(\uppercase\expandafter{\romannumeral1}) applying an anomaly detection method to the input image, 
(\uppercase\expandafter{\romannumeral2}) binarizing the anomaly map and cropping sub-images accordingly, 
(\uppercase\expandafter{\romannumeral3}) feeding these sub-images and their corresponding masks into our mask-guided ViT and getting probability output,
and (\uppercase\expandafter{\romannumeral4}) merging predictions from each sub-image to generate the final output.
Table~\ref{tab:inference_time} presents the per-image inference time for each step.
Note that the time taken for step (\uppercase\expandafter{\romannumeral1}) depends on the anomaly detection method and is independent of our AnomalyNCD.
The majority of the inference time is concentrated in the MEBin and image cropping, which account for over 80\% of the total runtime.
MEBin uses the official OpenCV library for connection component calculations,
resulting in the inability to use CUDA acceleration.

\begin{table}[!t]
    \centering
    \setlength{\tabcolsep}{2.0mm}
    \resizebox{1.0\linewidth}{!}{
    \begin{tabular}{ccccccc}
        \toprule
         & \multicolumn{3}{c}{MVTec AD}  & \multicolumn{3}{c}{MTD} \\ 
        \cmidrule(l){2-4} \cmidrule(l){5-7} 
        Methods & NMI & ARI & $F_1$ & NMI & ARI & $F_1$ \\   
        \midrule
        AC \citeapp{appWACV2023AC} & 0.711 & 0.638 & 0.718 & 0.467 & 0.359 & 0.482 \\
        Ours & \textbf{0.871} & \textbf{0.851} & \textbf{0.909} & \textbf{0.829} & \textbf{0.863} & \textbf{0.841} \\   
        \bottomrule
    \end{tabular}
    }
    \vspace{-0.5em}
    \caption{The quantitative results of using the ground truth masks. We compare our AnomalyNCD with Anomaly Clustering on the MVTec AD and MTD datasets.}
    \label{tab:gt_mask}
\end{table}

\section{Analysis of model performance using ground truth masks}
\label{app:gt_mask}
In the experimental section, we demonstrate that our multi-class classification performs better when the anomaly detection method performs better.
So in order to test the optimal results of our AnomalyNCD, we assume an ideal anomaly detection method where the ground truth masks of unlabeled images are available.
We report the results in Table~\ref{tab:gt_mask}, our approach achieves significant improvement over Anomaly Clustering on two datasets.
On MVTec AD, there is a 16.0\% improvement on the NMI and a 19.1\% improvement on the $F_1$, and on MTD, there is a 36.2\% improvement on the NMI and a 35.9\% improvement on the $F_1$.
In this ideal case, ground truth masks do not introduce any over-detections and missed detections compared to anomaly maps generated by anomaly detection methods.
With the improvement of anomaly detection methods in the future, our AnomalyNCD can achieve better results.

\begin{table*}[!ht]
    \centering
  \setlength{\tabcolsep}{4.0mm}
  \resizebox{\linewidth}{!}{
\begin{tabular}{c|cc|cc|cc|cc|c}
\hline
\textbf{PLC Thresholds} & \textbf{MVTec AD - MuSc} & \textbf{MTD - MuSc} & \textbf{MVTec AD - RD++} & \textbf{MTD - RD++} & \textbf{MVTec AD - PatchCore}& \textbf{MTD - PatchCore} & \textbf{Rank} \\
\hline
\textbf{0.1} & 0.602/0.522/0.715 & 0.212/0.180/0.419 & 0.625/0.531/0.717 & 0.368/0.377/0.587 &  0.630/0.574/0.734& 0.457/\textbf{0.453}/0.683 & 3.44\\
\hline
\textbf{0.3} & 0.615/0.533/0.727 & 0.230/0.204/0.487 & 0.627/0.545/0.721 & \textbf{0.391}/\textbf{0.382}/0.595 &  0.646/0.594/0.757& 0.466/0.452/0.683 & 2.28\\
\hline
\textbf{0.5} & 0.613/0.526/0.712 & \textbf{0.268}/\textbf{0.228}/\textbf{0.509} & 0.631/0.542/0.721 & 0.368/0.361/\textbf{0.600} & \textbf{0.670}/\textbf{0.601}/\textbf{0.769}& 0.380/0.390/\textbf{0.715} & \textbf{2.17}\\
\hline
\textbf{0.7} & \textbf{0.630}/\textbf{0.557}/\textbf{0.731} & 0.257/0.212/0.489 & \textbf{0.650}/\textbf{0.561}/\textbf{0.741} & 0.353/0.333/0.566 & 0.653/0.578/0.748& \textbf{0.602}/0.415/0.664 & 2.22\\
\hline
\textbf{0.9} & 0.584/0.476/0.682 & 0.231/0.172/0.489 & 0.604/0.487/0.682 & 0.337/0.302/0.549 & 0.506/0.338/0.610& 0.423/0.352/0.660 & 4.67\\
\hline
\end{tabular}}
\captionof{table}{The results (\textbf{NMI}, \textbf{ARI}, \textbf{F1}) on two datasets, when taking various PLC thresholds.
\textbf{Rank} is the average ranking of the threshold.}
\label{tab:plc}
\end{table*}

\begin{table}[!t]
    \centering
    \setlength{\tabcolsep}{2.0mm}
    \resizebox{1.0\linewidth}{!}{
    \begin{tabular}{ccccccc}
        \toprule
         & \multicolumn{3}{c}{MuSc\citeapp{appICLR2024MuSc}+AnomalyNCD}  & \multicolumn{3}{c}{CPR\citeapp{appTIP2024CPR}+AnomalyNCD} \\ 
        \cmidrule(l){2-4} \cmidrule(l){5-7}
        & NMI & ARI & $F_1$ & NMI & ARI & $F_1$ \\   
        \midrule
        $\tau=2$ & 0.640 & 0.537 & 0.707 & 0.720 & 0.647 & 0.787 \\
        $\tau=3$ & 0.647 & 0.543 & 0.718 & 0.736 & 0.668 & 0.797 \\ 
        $\tau=4$ & 0.613 & 0.526 & 0.712 & 0.736 & 0.674 & 0.805 \\ 
        $\tau=5$ & 0.618 & 0.539 & 0.740 & 0.721 & 0.665 & 0.807 \\ 
        $\tau=6$ & 0.600 & 0.510 & 0.715 & 0.710 & 0.655 & 0.799 \\
        \midrule
        variance & 0.0003 & 0.0001 & 0.0001 & 0.0001 & 0.0001 & 0.0001 \\
        \bottomrule
    \end{tabular}
    }
    \vspace{-0.5em}
    \caption{Sensitivity analysis of the minimum stable range $\tau$ in MEBin. We use the zero-shot AD method MuSc and one-class AD method CPR to conduct experiments.}
    \label{tab:min_stable}
    \vspace{-0.5em}
\end{table}

\begin{table*}[!b]
    \centering
    \resizebox{1.0\linewidth}{!}{
    \begin{tabular}{c|ccc|ccc|ccc|ccc|ccc|ccc}
        \toprule
        \multirow{2}{*}{\textbf{Methods}} & \multicolumn{3}{c}{\textbf{MuSc\citeapp{appICLR2024MuSc}}} & \multicolumn{3}{c}{\textbf{PatchCore\citeapp{appCVPR2022patchcore}}} & \multicolumn{3}{c}{\textbf{EfficientAD\citeapp{appwacv2024efficientad}}} & \multicolumn{3}{c}{\textbf{RD++\citeapp{appcvpr2023rd}}} & \multicolumn{3}{c}{\textbf{PNI\citeapp{appiccv2023pni}}} & \multicolumn{3}{c}{\textbf{CPR\citeapp{appTIP2024CPR}}} \\
         & \multicolumn{3}{c}{\textbf{+AnomalyNCD}} & \multicolumn{3}{c}{\textbf{+AnomalyNCD}} & \multicolumn{3}{c}{\textbf{+AnomalyNCD}} & \multicolumn{3}{c}{\textbf{+AnomalyNCD}} & \multicolumn{3}{c}{\textbf{+AnomalyNCD}} & \multicolumn{3}{c}{\textbf{+AnomalyNCD}} \\ \hline
        \textbf{Products$\backslash$Metric} & NMI & ARI & $F_1$ & NMI & ARI & $F_1$ & NMI & ARI & $F_1$ & NMI & ARI & $F_1$ & NMI & ARI & $F_1$ & NMI & ARI & $F_1$  \\ 
        \midrule
        bottle & 0.613 & 0.583 & 0.819 & 0.734 & 0.671 & 0.855 & 0.544 & 0.467 & 0.759 & 0.640 & 0.529 & 0.771 & 0.676 & 0.598 & 0.819 & 0.775 & 0.757 & 0.904  \\ 
        cable & 0.597 & 0.492 & 0.626 & 0.758 & 0.666 & 0.741 & 0.549 & 0.414 & 0.619 & 0.679 & 0.602 & 0.734 & 0.765 & 0.601 & 0.691 & 0.711 & 0.476 & 0.561  \\ 
        capsule & 0.445 & 0.335 & 0.591 & 0.402 & 0.277 & 0.530 & 0.519 & 0.402 & 0.644 & 0.481 & 0.392 & 0.606 & 0.425 & 0.338 & 0.568 & 0.549 & 0.438 & 0.674  \\ 
        carpet & 0.852 & 0.837 & 0.906 & 0.628 & 0.569 & 0.726 & 0.672 & 0.577 & 0.726 & 0.599 & 0.527 & 0.658 & 0.619 & 0.548 & 0.701 & 0.740 & 0.638 & 0.795  \\ 
        grid & 0.622 & 0.578 & 0.731 & 0.670 & 0.581 & 0.731 & 0.684 & 0.570 & 0.795 & 0.631 & 0.544 & 0.615 & 0.583 & 0.534 & 0.705 & 0.766 & 0.689 & 0.731  \\ 
        hazelnut & 0.662 & 0.582 & 0.718 & 0.859 & 0.875 & 0.927 & 0.570 & 0.429 & 0.673 & 0.845 & 0.853 & 0.909 & 0.919 & 0.926 & 0.955 & 0.723 & 0.727 & 0.827  \\ 
        leather & 0.863 & 0.838 & 0.911 & 0.770 & 0.734 & 0.839 & 0.587 & 0.449 & 0.621 & 0.746 & 0.688 & 0.782 & 0.672 & 0.616 & 0.734 & 0.865 & 0.827 & 0.895  \\ 
        metal\_nut & 0.643 & 0.467 & 0.565 & 0.910 & 0.883 & 0.948 & 0.436 & 0.272 & 0.443 & 0.851 & 0.821 & 0.922 & 0.891 & 0.872 & 0.948 & 0.870 & 0.848 & 0.930  \\ 
        pill & 0.439 & 0.291 & 0.513 & 0.476 & 0.335 & 0.567 & 0.388 & 0.194 & 0.460 & 0.467 & 0.286 & 0.527 & 0.461 & 0.315 & 0.580 & 0.694 & 0.592 & 0.747  \\ 
        screw & 0.399 & 0.265 & 0.488 & 0.638 & 0.569 & 0.756 & 0.594 & 0.498 & 0.719 & 0.615 & 0.551 & 0.725 & 0.640 & 0.616 & 0.769 & 0.742 & 0.698 & 0.800  \\ 
        tile & 0.885 & 0.850 & 0.940 & 0.927 & 0.912 & 0.966 & 0.523 & 0.375 & 0.598 & 0.757 & 0.711 & 0.795 & 1.000 & 1.000 & 1.000 & 0.98 & 0.976 & 0.992  \\ 
        toothbrush & 0.368 & 0.259 & 0.762 & 0.271 & 0.084 & 0.667 & 0.398 & 0.313 & 0.786 & 0.316 & 0.164 & 0.714 & 0.271 & 0.084 & 0.667 & 0.508 & 0.499 & 0.857  \\ 
        transistor & 0.531 & 0.421 & 0.620 & 0.642 & 0.693 & 0.760 & 0.439 & 0.462 & 0.650 & 0.477 & 0.292 & 0.500 & 0.716 & 0.743 & 0.820 & 0.577 & 0.523 & 0.700  \\ 
        wood & 0.743 & 0.672 & 0.868 & 0.782 & 0.644 & 0.868 & 0.375 & 0.196 & 0.574 & 0.750 & 0.651 & 0.868 & 0.850 & 0.790 & 0.927 & 0.816 & 0.756 & 0.912  \\ 
        zipper & 0.526 & 0.417 & 0.615 & 0.583 & 0.518 & 0.652 & 0.455 & 0.291 & 0.548 & 0.608 & 0.515 & 0.689 & 0.639 & 0.552 & 0.659 & 0.726 & 0.667 & 0.756  \\ 
        \hline
        \textbf{Mean} & 0.613 & 0.526 & 0.712 & 0.670 & 0.601 & 0.769 & 0.516 & 0.394 & 0.641 & 0.631 & 0.542 & 0.721 & 0.675 & 0.609 & 0.769 & 0.736 & 0.674 & 0.805  \\ 
        \bottomrule
    \end{tabular}
    }
    \caption{Detailed quantitative results on the MVTec AD dataset.}
    \label{tab:detailed_mvtec}
\end{table*}

\section{Ablation study on the number of threshold index in MEBin}
\label{app:sampling}
In Fig.~\ref{fig:mebin}, we conduct the ablation study on the number of threshold index $\mathcal{T}$.
Note that we use powers of 2 for candidates of $\mathcal{T}$, since it is calculated on grayscale images (0-255).
$\mathcal{T}\!=\!64 $ achieves better FPR-FNR trade-off compared to other $\mathcal{T}$. 
With a small value ($\mathcal{T}=32$), the binary mask changes dramatically, making it difficult to determine a stable connected component.
Conversely, a larger $\mathcal{T}$ brings more time consumption.
Therefore, we set $\mathcal{T}=64$ as the default value.

\section{Sensitivity analysis of $\tau$ in MEBin}
\label{app:min_stable}
In Table~\ref{tab:min_stable}, we experiment with the parameter sensitivity of the minimum stable range $\tau$ on the MVTec AD dataset.
We use the zero-shot anomaly detection method MuSc \citeapp{appICLR2024MuSc} and the one-class method CPR \citeapp{appTIP2024CPR} respectively to combine with our AnomalyNCD.
When the hyperparameter $\tau$ is changed, the multi-class anomaly classification results change with a variance of about 0.0001, demonstrating that our method is insensitive to $\tau$.

\section{Sensitivity analysis of threshold 0.5 in PLC}
\label{app:plc}
Although MEBin achieves a decent trade-off,
it may still get over-detections due to local noises. 
Thus, PLC uses the threshold $0.5$ to find over-detections 
and corrects their pseudo label with a normal one-hot label.
The optimal PLC threshold is related to datasets and anomaly detection methods.
Tab.~\ref{tab:plc} shows the results of three methods on two datasets, when taking various PLC thresholds.
Observably, a threshold of 0.5 ranks $\text{1}^\text{st}$ performance on average in all candidates.

\section{Binarization results of MEBin}
\label{app:bin_map}
\begin{figure*}[h]
\centering
\includegraphics[width=0.9\textwidth]{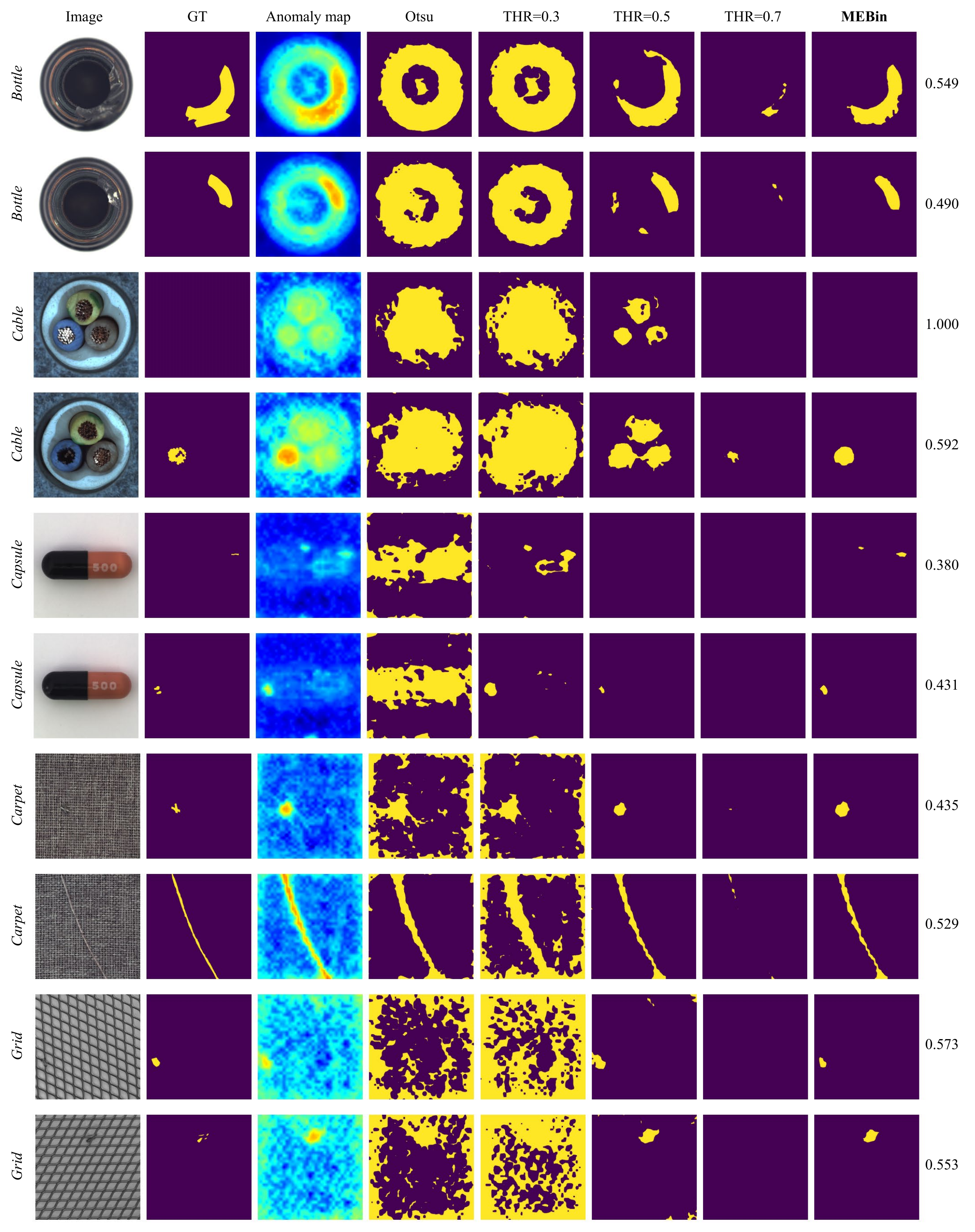}
\caption{
\textbf{Binarization results of MEBin on the MVTec AD dataset.} We show for each category: RGB image, ground truth, anomaly map, binary map of Otsu method, binary map of fixed threshold 0.3, binary map of fixed threshold 0.5, binary map of fixed threshold 0.7, and binary map of our MEBin.
We report the optimal threshold searched by our MEBin in the last column.
}
\label{fig:mebin_vis_1}
\end{figure*}

\begin{figure*}[h]
\centering
\includegraphics[width=0.9\textwidth]{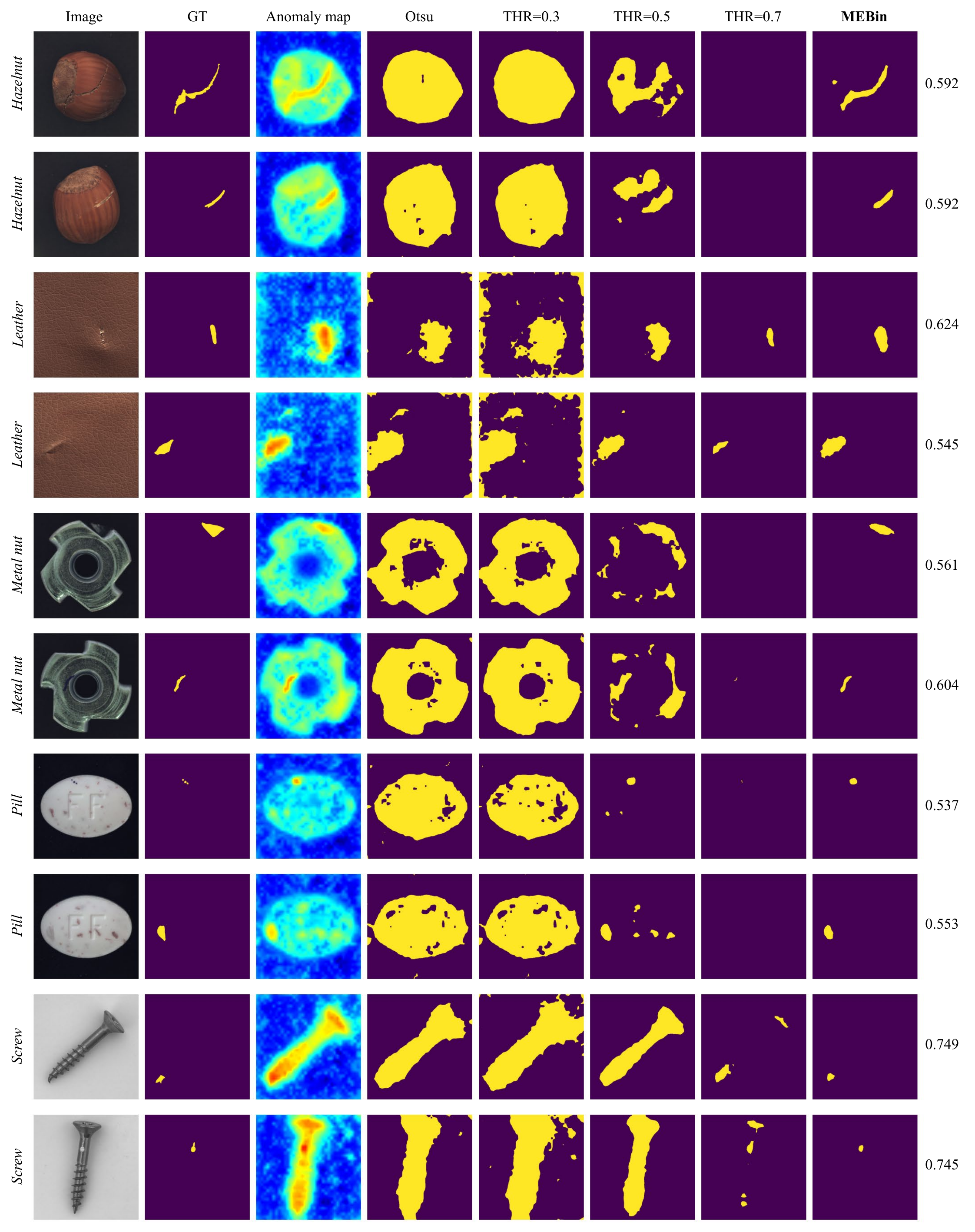}
\caption{
\textbf{Binarization results of MEBin on the MVTec AD dataset.} We show for each category: RGB image, ground truth, anomaly map, binary map of Otsu method, binary map of fixed threshold 0.3, binary map of fixed threshold 0.5, binary map of fixed threshold 0.7, and binary map of our MEBin.
We report the optimal threshold searched by our MEBin in the last column.
}
\label{fig:mebin_vis_2}
\end{figure*}

\begin{figure*}[h]
\centering
\includegraphics[width=0.9\textwidth]{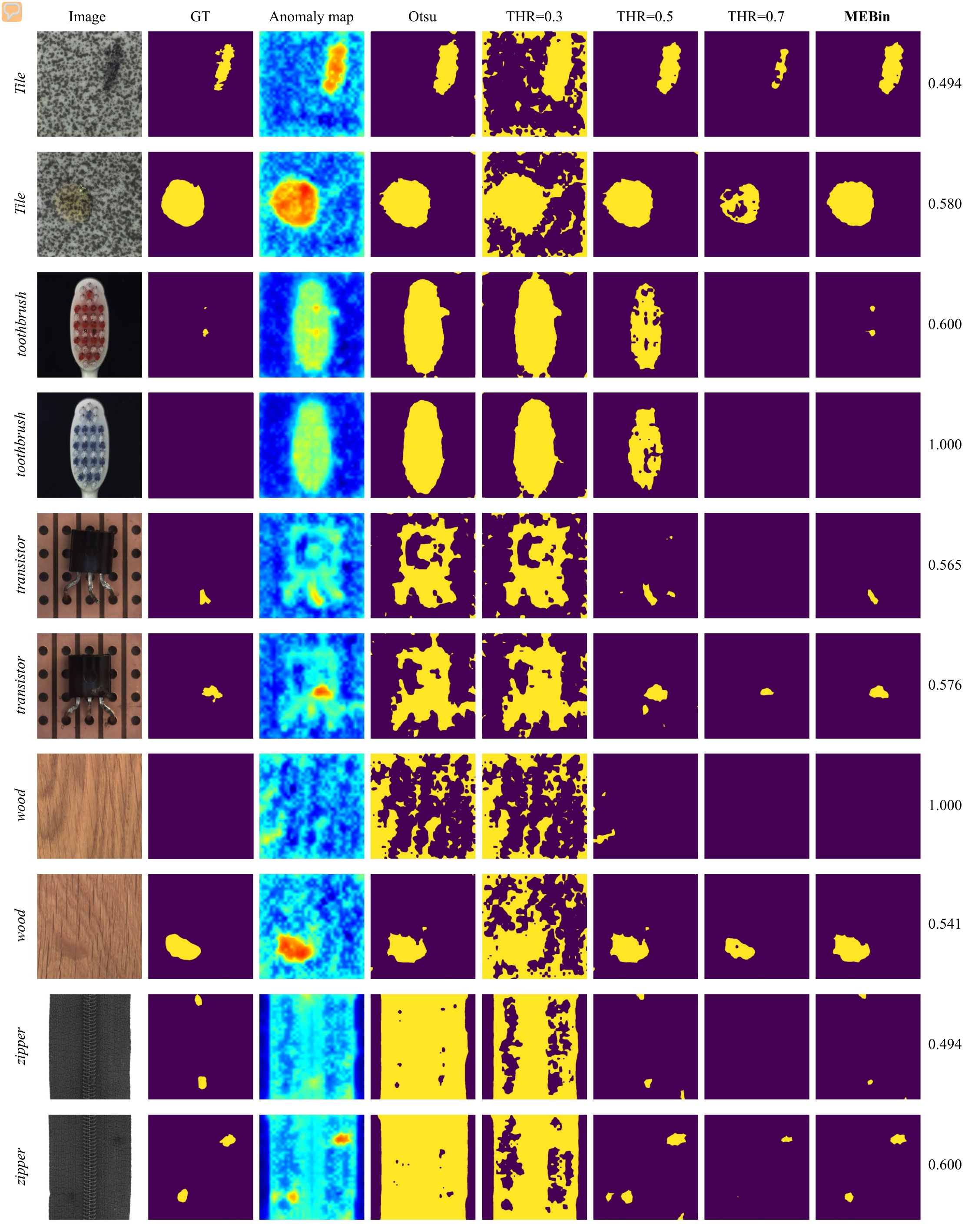}
\caption{
\textbf{Binarization results of MEBin on the MVTec AD dataset.} We show for each category: RGB image, ground truth, anomaly map, binary map of Otsu method, binary map of fixed threshold 0.3, binary map of fixed threshold 0.5, binary map of fixed threshold 0.7, and binary map of our MEBin.
We report the optimal threshold searched by our MEBin in the last column.
}
\label{fig:mebin_vis_3}
\end{figure*}

From Fig.~\ref{fig:mebin_vis_1} to Fig.~\ref{fig:mebin_vis_3}, we binarize the anomaly maps output by MuSc \citeapp{appICLR2024MuSc} using different binarization strategies, including the fixed threshold, the Otsu \citeapp{appTSMC1979otsu} method, and our MEBin.
Our binarization strategy does not require the validation set to determine a threshold
and adaptively obtain the optimal threshold for each image with fewer false positives and false negatives.
Compared to the Otsu method, our MEBin has fewer false positives, especially on normal images.
We also report the optimal threshold searched by our MEBin in the last column of Fig.~\ref{fig:mebin_vis_1} to Fig.~\ref{fig:mebin_vis_3}.
The optimal threshold ranges from 0.380 to 1.000, making it difficult to handle all situations with a fixed threshold.

\begin{figure}[!t]
\centering
\includegraphics[width=0.48\textwidth]{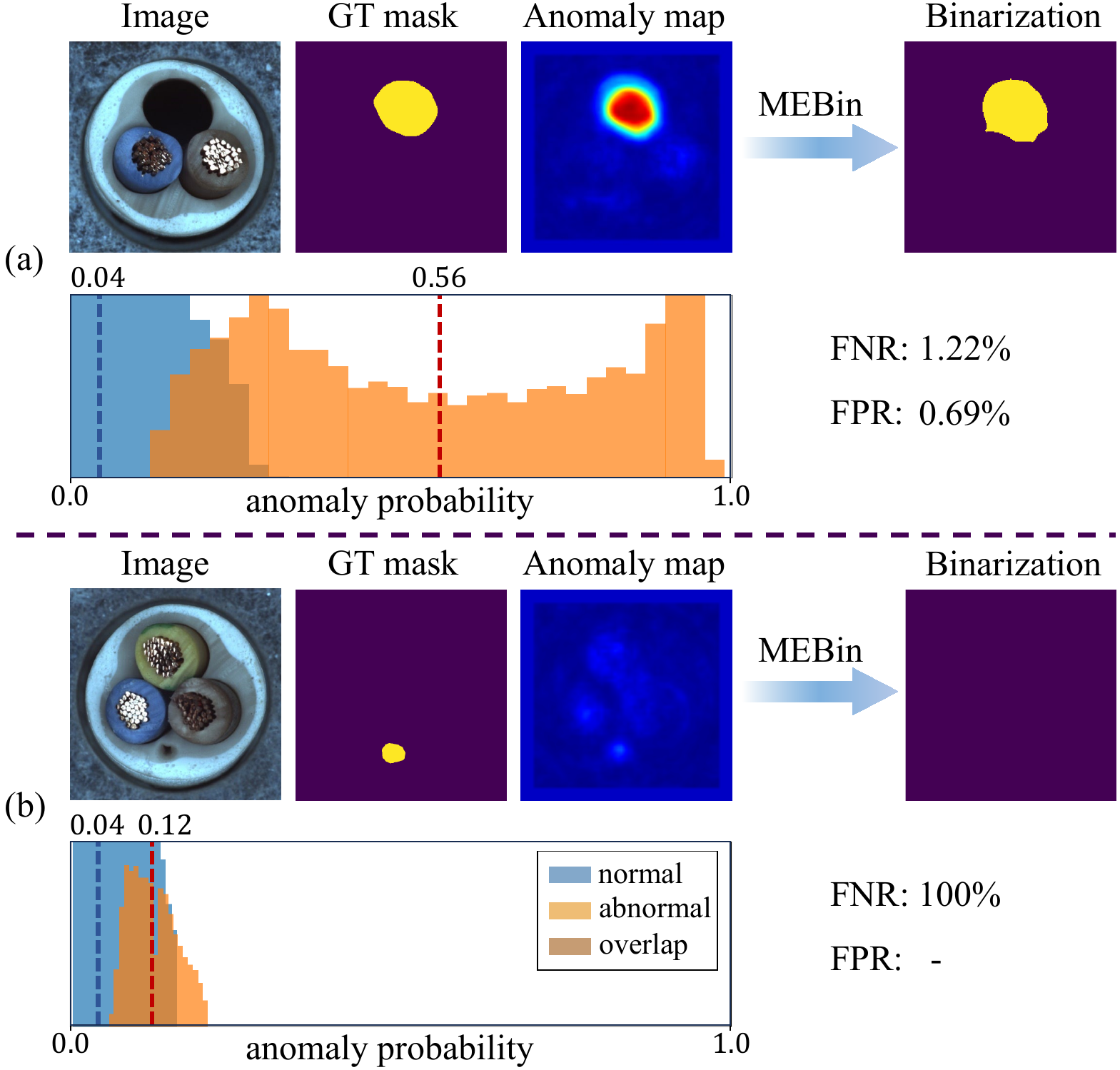}
\caption{\textbf{Analysis of anomaly maps in EfficientAD.} We visualize the pixel-level anomaly probability histogram, and use dashed lines to represent the average anomaly probability for normal and anomalous regions respectively.}
\vspace{-0.5em}
\label{efficient_ad}
\end{figure}

\section{Detailed t-SNE visualization}
\label{app:detailed_tsne}
\begin{figure*}[t]
\centering
\includegraphics[width=0.95\textwidth]{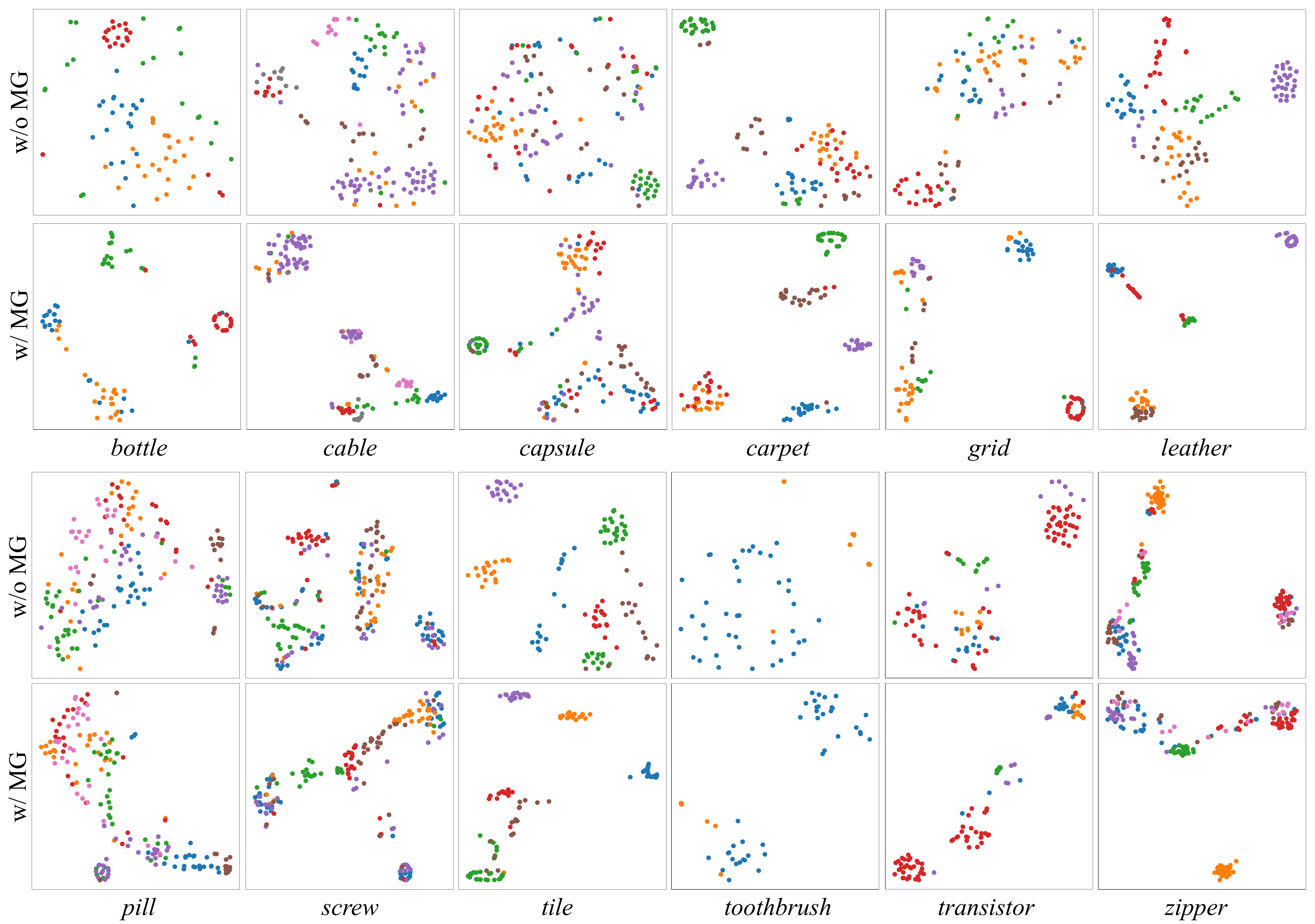}
\caption{
\textbf{T-SNE visualization of sub-images on the MVTec AD dataset.} The different colors of dots represent their anomaly classes.
}
\label{fig:tsne_all}
\end{figure*}

\begin{figure*}[b]
\centering
\includegraphics[width=0.95\textwidth]{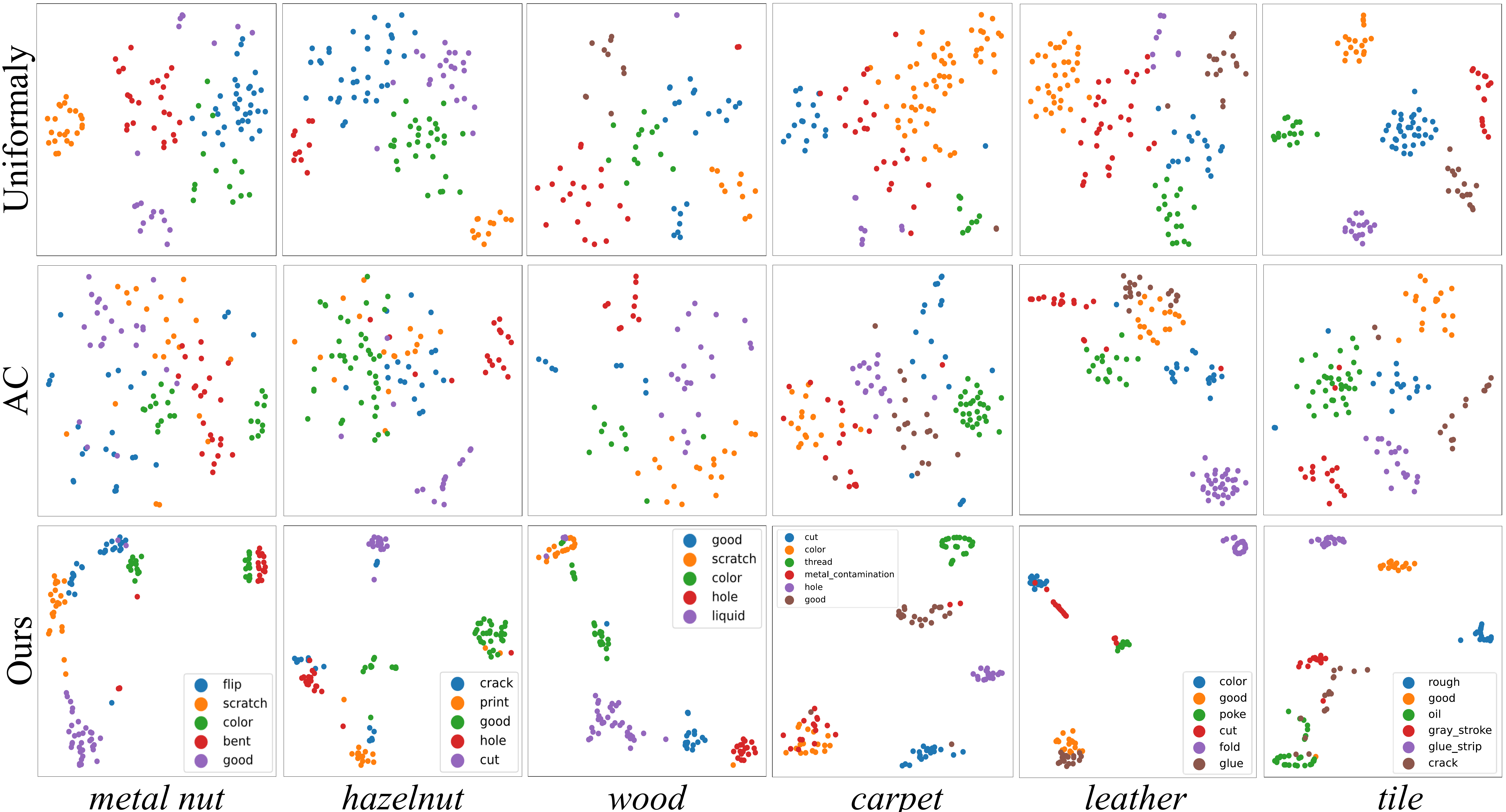}
\caption{Qualitative comparisons with different clustering methods by t-SNE visualization.}
\label{fig:tsne_compare}
\end{figure*}

In Fig.~\ref{fig:tsne_all}, we show more t-SNE visualization on all the products of the MVTec AD dataset.
We also provide more qualitative comparisons with AC \citeapp{appWACV2023AC} and Uniformaly \citeapp{appArxiv2023uniformaly} in Fig.~\ref{fig:tsne_compare}.
These methods show slight clustering phenomena, while our AnomalyNCD has larger inter-class distances and smaller intra-class distances.
This further proves the superiority of our method from the qualitative perspective.

\section{Detailed quantitative results}
\label{app:detailed_mvtec}

In this section, we report the detailed results of our AnomalyNCD combined with various anomaly detection methods on the MVTec AD dataset, 
as shown in Table~\ref{tab:detailed_mvtec}.

\section{Limitation}
\label{app:limit}
The anomaly detection methods used affect the performance of our AnomalyNCD.
In general, if the anomaly detection method has a higher AUPRO, we can achieve a better performance in the multi-class anomaly classification task.
In the ideal situation where the ground truth mask is available (AUPRO=100\%), our AnomalyNCD is significantly superior to other methods, which has been discussed in Sec.~\ref{app:gt_mask}.
However, EfficientAD \citeapp{appwacv2024efficientad} is an exception, with a lower multi-class anomaly classification performance despite higher AUPRO.
We observe that a large span of anomaly probability values exists between different anomaly maps generated by EfficientAD, as shown in Fig.~\ref{efficient_ad}.
The average value of the anomalous region in (a) is only 0.12, which is close to the anomaly probability in the normal region, while that in (b) is 0.56.
This large span causes false negatives in (a) during binarization, resulting in a lower performance for anomaly classification.

\clearpage

{
    \small
    \bibliographystyleapp{ieeenat_fullname}
    \balance
    \bibliographyapp{suppl}
}


\end{document}